\documentclass[runningheads]{llncs}
\usepackage{graphicx}

\usepackage{tikz}
\usepackage{comment}
\usepackage{adjustbox}
\usepackage{amsmath,amssymb} 
\usepackage{color}
\usepackage{multirow}
\usepackage{subcaption}
\usepackage{pifont}
\newcommand{\cmark}{\ding{51}}%
\newcommand{\xmark}{\ding{55}}%
\usepackage[hidelinks]{hyperref}

\usepackage[accsupp]{axessibility}  

\usepackage{listings}
\usepackage{xcolor}

\definecolor{codegreen}{rgb}{0,0.6,0}
\definecolor{codegray}{rgb}{0.5,0.5,0.5}
\definecolor{codepurple}{rgb}{0.58,0,0.82}
\definecolor{backcolour}{rgb}{0.95,0.95,0.92}

\lstdefinestyle{mystyle}{
    backgroundcolor=\color{backcolour},   
    commentstyle=\color{codegreen},
    keywordstyle=\color{magenta},
    numberstyle=\tiny\color{codegray},
    stringstyle=\color{codepurple},
    basicstyle=\ttfamily\footnotesize,
    breakatwhitespace=false,         
    breaklines=true,                 
    captionpos=b,                    
    keepspaces=true,                 
    numbers=left,                    
    numbersep=5pt,                  
    showspaces=false,                
    showstringspaces=false,
    showtabs=false,                  
    tabsize=2
}

\begin{document}
\pagestyle{headings}
\mainmatter
\def\ECCVSubNumber{7149}  

\title{Improving Semantic Segmentation in Transformers using Hierarchical Inter-Level Attention} 


\titlerunning{Hierarchical Inter-Level Attention}
\author{%
Gary Leung $^{1,3}$  \and Jun Gao$^{1,2,3}$   \and Xiaohui Zeng $^{1,3}$  \and  Sanja Fidler$^{1,2,3}$ \vspace{8pt}\\
\small{University of Toronto\textsuperscript{1} \quad Nvidia\textsuperscript{2} \quad Vector Institute\textsuperscript{3} } \vspace{3pt}\\
\texttt{\scriptsize \{garyleung,jungao,xiaohui,fidler\}@cs.toronto.edu}
}
\authorrunning{Leung et al.}
\institute{}

\newcommand{\methodfull}{{\color{purple}Hierarchical Inter-Level Attention}}
\newcommand{\method}{{\color{purple}HILA}}
\newcommand{\methodone}[1]{{\color{purple}HILA S(3)}}
\newcommand{\methodthree}[1]{{\color{purple}HILA S(2,3,4)}}
\newcommand{\uncertain}[1]{{#1}}
\newcommand{\todo}[1]{{\color{blue}TODO: #1}}

\newcommand{\gary}[1]{{\color{blue}{[Gary: #1]}}}
\newcommand{\JG}[1]{{\color{orange}{[Jun: #1]}}}
\newcommand{\SF}[1]{{\color{magenta}{[Sanja: #1]}}}
\newcommand{\xh}[1]{{\color{teal}{[xiaohui: #1]}}}
\maketitle

\begin{abstract}
Existing transformer-based 
image backbones 
typically propagate feature information in one direction from lower to higher-levels. This may not be ideal since the localization ability to delineate accurate object boundaries, is most prominent in the lower, high-resolution feature maps, while the semantics that can disambiguate image signals belonging to one object 
vs.
another, typically emerges in a higher level of processing. 
We present \methodfull{} (\method{}), an attention-based method that captures Bottom-Up and Top-Down Updates between features of different levels. \method{} extends hierarchical vision transformer architectures by adding local connections between features of higher and lower levels to the backbone encoder. In each iteration, we construct a hierarchy by having higher-level features compete for assignments to update lower-level features belonging to them, iteratively resolving object-part relationships. These improved lower-level features are then used to re-update the higher-level features. \method{} can be integrated into the majority of hierarchical architectures without requiring any changes to the base model. We add \method{} into SegFormer and the Swin Transformer and show notable improvements in accuracy 
in semantic segmentation with fewer parameters and FLOPS. Project website and code: \url{cs.toronto.edu/~garyleung/hila}.


\end{abstract}

\vspace{-7mm}
\section{Introduction}
\vspace{-2mm}
Semantic segmentation is one of the fundamental  tasks in computer vision, with several downstream applications such as autonomous driving \cite{siam2017deepauto,teichmann123autonomous}, medical imaging~\cite{ronneberger2015unet}, and conditional image synthesis~\cite{park2019SPADE}. Traditionally, Convolutional Neural Networks (CNNs) \cite{lecun1995convolutional,long2015fcn,takikawa2019gatedscnn,chen2018deeplabv3+} have been used with great success in major segmentation benchmarks. More recently, following the success of Transformers \cite{vaswani2017attention} in Natural Language Processing, significant interest has been seen in leveraging transformer-based models for vision related tasks~\cite{dosovitskiy2020image}. Different from CNNs, Hierarchical Vision Transformers (HVT) \cite{hvt:xie2021segformer,hvt:liu2021Swin,hvt:wang2021pvtv2,hvt:yang2021focal,hvt:chu2021twins,hvt:dong2021cswin,hvt:wu2021pale} utilize self-attention, allowing for global content-dependent interactions amongst features within the same resolution level. Recent HVT-based models have surpassed CNNs and achieved state-of-the-art on several vision tasks such as image classification \cite{hvt:wu2021pale} and semantic segmentation \cite{hvt:xie2021segformer,hvt:liu2021Swin}.


Current architectures \cite{hvt:xie2021segformer,chen2018deeplabv3+,huang2019ccnet,fu2019danet,cheng2021maskformer} use a backbone encoder to create the initial hierarchical feature map, before processing these features into the segmentation output. One issue with common backbone encoders such as ResNet \cite{he2016resnet}, or recent vision transformer-based encoders \cite{hvt:liu2021Swin,hvt:xie2021segformer,hvt:chu2021twins,hvt:yang2021focal,hvt:wang2021pvtv2} is that features at different levels are generated sequentially, with higher-level features building on top of lower-level features. We argue that this may not be ideal since the localization ability, such as delineating object boundaries or object parts in an image, is most prominent in the lower, high-resolution feature maps, while the semantics that can disambiguate image signals belonging to one object vs another, typically only emerges in higher levels of processing~\cite{zeiler2014visualizing,bau2017network}. 


Several architectures such as DLA~\cite{yu2018dla}, DenseNet~\cite{huang2017densenet}, UNet~\cite{ronneberger2015unet} and FPN-based models~\cite{zhao2021graphfpn,zhang2020fpt,lin2019zigzagnet} fuse information from multiple levels in order to preserve both high-level semantic information and low-level high-frequency details. However, while fusion is certainly helpful in predicting more detailed and localized semantic outputs, it may not be the best solution to resolve ambiguities in the hierarchical processing chain. For example, in the region of occlusion where two or more objects co-locate, if features at early levels mix information from all co-located objects, these ambiguities may propagate upward in the hierarchy. Fusing lower level features with higher level features may thus lead to spurious boundaries. On the other hand, in architectures such as Capsules~\cite{sabour2017dynamic}, ambiguities get resolved as higher level ``object" capsules compete to explain lower level ``part" capsules. We take inspiration from this work.

We propose \methodfull{} (\method{}), an attention-based method that adds both Top-Down and Bottom-Up Interactions between features of different Levels. We extend Hierarchical Vision Transformer (HVT) backbones by adding two updates to pre-existing blocks. In our Top-Down Update, higher-level features compete to improve lower-level features using our Inter-Level Attention. High-level information is propagated downwards in a local patch area, resolving ambiguities in lower-level features and improving their semantics. Our Bottom-Up Update then uses our Inter-Level Attention to selectively propagate the improved lower-level features back upwards, improving the semantic clarity of our higher-level features. Our attention update is computationally lightweight and can be integrated into the majority of existing HVT architectures without requiring any changes to the base model. 

We demonstrate the effectiveness of \method{} by applying it to state-of-the-art transformer models SegFormer \cite{hvt:xie2021segformer} and Swin Transformer \cite{hvt:liu2021Swin}. Our results show notable improvements in 
accuracy
on two public datasets: Cityscapes \cite{cordts2016cityscapes} and ADE20K \cite{ade20k}. Notably, the Segformer B1 model augmented with \methodthree{} 
improves over the baseline Segformer-B1 model performance by +2.4 mIOU /+3.4 F-Score and surpasses Segformer-B2 model's performance with 21\% less params and 41\% less FLOPS. In real-time performance tests, models augmented with \method{} can achieve similar mIOU and F-Score while having 126\% more FPS. Moreover, \method{} creates  interpretable hierarchical visualization results which aid in understanding the behaviour of higher-level features.

\begin{figure}[t!]
\centering
\vspace{-2mm}
\includegraphics[width=0.95\linewidth]{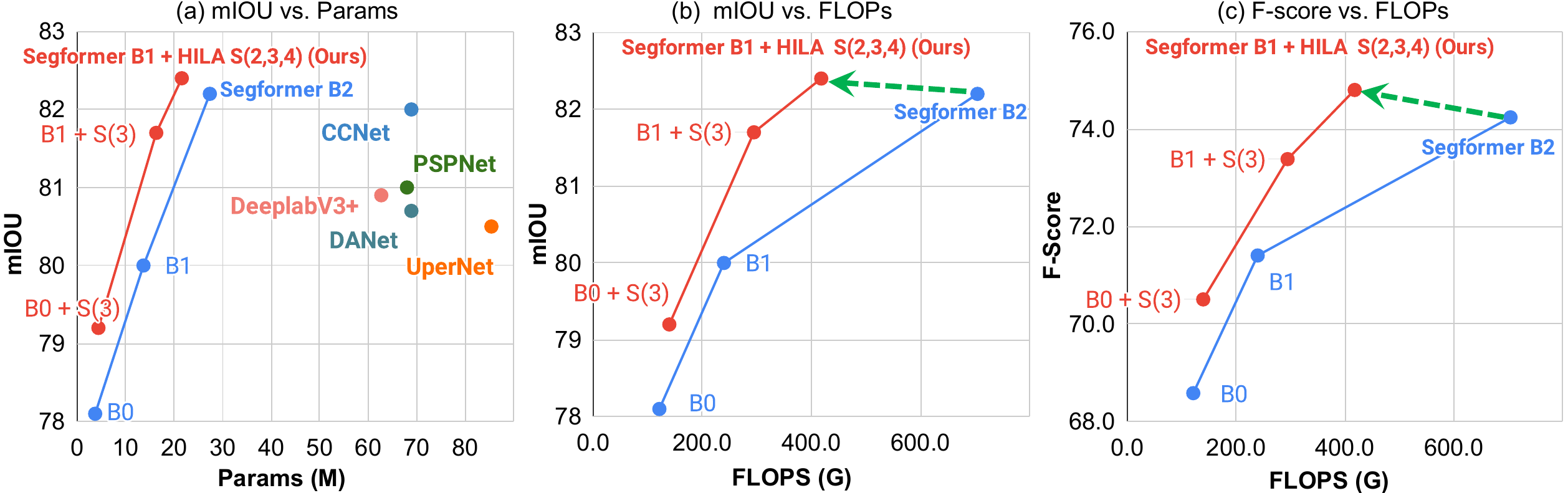}
\vspace{-4mm} 
\caption{\footnotesize \textbf{Performance vs. Model Efficiency on Cityscapes.} mIoU evaluated using multi-scale inference and F-score (3px threshold) using single-scale inference. S() refers to the backbone stages that HILA is applied to. We show significant improvements from  adding \method{} to the state-of-the-art Segformer model. 
}
\label{fig:performancegraph}
\vspace{-6mm}
\end{figure}

\vspace{-3mm}
\section{Related Work}
\vspace{-2mm}
\vspace{-3mm}
\paragraph{Semantic Segmentation.} Fully Convolutional Neural Networks (FCN) \cite{long2015fcn} is one of the seminal works that leveraged CNNs for the task of semantic segmentation in an end-to-end manner. Many follow-up methods looked to improve FCN, such as in increasing the model's receptive field~\cite{yu2015multi,chen2018deeplabv3+}, in integrating boundary information~\cite{takikawa2019gated}, and in leveraging attention into modules \cite{wang2018non}. 
More recently, there has been significant interest in integrating transformers to improve segmentation architectures. Methods like MaskFormer \cite{cheng2021maskformer} integrate self-attention to aid in processing backbone features. On the other hand, methods like 
SegFormer \cite{hvt:xie2021segformer} replace the backbone encoder entirely with a hierarchical vision transformer.
\method{} builds on the state-of-the-art HVT backbones with inter-level connections to iteratively propagate information between levels when extracting features. 

\vspace{-3mm}
\paragraph{Vision Transformers.} Vision Transformer (ViT) \cite{dosovitskiy2020vit} is the seminal work to demonstrate the effectiveness of Transformer Encoders in the image classification task. It divides the images into sequences of tokens which are then processed with self-attention alongside a \textsc{class} token for the image category prediction. However, since ViT maintains one fixed resolution of the feature map, it is not
computationally efficient for dense prediction tasks, especially for semantic segmentation for large images.
The hierarchical vision transformer (HVT) architecture was further introduced by PVT \cite{hvt:wang2021pyramid} and Swin \cite{hvt:liu2021Swin} to alleviate this issue.  PVT and Swin used patch merging to create hierarchical feature representations and proposed spatial-reduction attention~\cite{hvt:wang2021pyramid}  or local shifted window~\cite{hvt:liu2021Swin} to make self-attention computationally efficient at higher resolutions. Follow up work \cite{hvt:chu2021twins,hvt:wang2021pvtv2,hvt:yang2021focal,ffn:li2021localvit,hvt:wu2021pale,hvt:dong2021cswin,hvt:chen2021dpt} largely uses the same architecture structure and focuses on enhancing the efficiency and expressiveness of self-attention. Specifically, Twins \cite{hvt:chu2021twins} integrates local window attention and global pooling attention, Focal Transformer \cite{hvt:yang2021focal} introduces focal self-attention where the granularity of attention changes based on distance, and Pyramid Pooling Transformer~\cite{hvt:wu2021p2t} utilizes pyramid pooling to obtain efficient representations while capturing contextual features. One common limitation amongst HVT works is that features are generated sequentially, without any inter-level connections beyond the patch merging step. \method{} changes this by integrating iterative inter-level connections into pre-existing HVT works to gradually 
generate the feature hierarchy. In work parallel to ours, HRViT~\cite{gu2022hrvit} leverages convolutional inter-level connections in the patch-embedding step of a specially structured HVT. \method{} differs from HRViT in that our inter-level connections leverage attention and occur iteratively multiple times in the transformer encoding step instead. \method{} is also applied to HVTs in a manner that is agnostic to different encoder backbones, unlike HRViT. 
\vspace{-3mm}
\paragraph{Inter-level Connections.} Several works aim to propagate information across features of different levels. Encoders like Deep Layer Aggregation (DLA)~\cite{yu2018dla}, DenseNet \cite{huang2017densenet} and more recently, D3Net~\cite{takahashi2021d3net} all have dense connections in the backbone encoder, propagating features from lower-levels to higher-levels. However, in these architectures, the features are always propagating in one direction, and the higher-level features process the potentially noisy and ambiguous lower-level features.
HRNet~\cite{sun2019hrnet} and HRViT~\cite{gu2022hrvit} introduces an encoder with convolutions connections in both directions through a singular fusion module. Different from HRNet and HRViT, \method{}'s inter-level connections use attention and occur iteratively multiple times, allowing an hierarchy to emerge and be refined gradually instead of directly outputted in one step such as in the fusion module. We go into more detail regarding differences in Section~\ref{differencepastwork}.
UNet~\cite{ronneberger2015unet} creates a hierarchical feature map, and then fuses it via propagating higher-level features into lower-level features with skip connections.
Feature Pyramid Network (FPN) \cite{lin2017fpn} processes the hierarchical feature outputs by adding top-to-bottom connections, propagating features from higher-level to lower-level to obtain rich semantics at all levels. Several recent works follow up on FPNs \cite{lin2019zigzagnet,liu2018panet,zhao2021graphfpn,zhang2020fpt}, with notable examples being GraphFPN \cite{zhao2021graphfpn}, which divides the features into a superpixel hierarchy and build inter-level connections, and Feature Pyramid Transformer (FPT) \cite{zhang2020fpt} which leverages attention for both top-down and bottom-up updates in the feature pyramid. 
While both FPT and GraphFPN are closely related to \method{} in terms of the attention updates between levels, the major difference is that \method{} updates the features at the current level before propagating information to the next level.  
This allows improved features to directly serve as better initialization for higher feature levels. Note that both \method{} and FPN  based methods \cite{lin2019zigzagnet,liu2018panet,zhao2021graphfpn,zhang2020fpt,lin2017fpn} 
can be used simultaneously as they affect different parts of the architecture.

Another line of research focuses on extracting a part-whole hierarchy from images. Capsules~\cite{sabour2017dynamic,hinton2018matrix} use an iterative EM algorithm to disambiguate the assignment between high-level capsules and low-level features. 
More recently, GLOM~\cite{hinton2021represent} proposed extracting the part-whole hierarchy by iteratively refining the features across all levels using bottom-up and top-down interactions. 
Visual Parser~\cite{bai2021visual} separates features into part-level and whole-level features which then iteratively updates each other. However this updating only occurs within the same level. \method{}'s design takes inspiration from these papers, and we design our Top-Down and Bottom-Up Updates to iteratively refine features and propagate information across different levels.
\vspace{-4mm}
\section{Method}
\vspace{-4mm}
\methodfull{} (\method{}) extends hierarchical vision transformer (HVT) by adding local attention-based updates between features of different levels. We follow common HVT terminology \cite{hvt:liu2021Swin,hvt:wang2021pvtv2} and use the term `stage' to refer to distinct levels in the architecture's hierarchy. Typically, HVT architectures process the input image sequentially through multiple stages and progressively outputs higher-level features at smaller resolutions. The higher level will normally have no feedback to the lower level, which potentially can extract noisy and incorrect features. \method{} changes this by introducing an Inter-Level Attention update into the feature encoder, 
allowing for feedback to propagate between different stages. 
We briefly describe the general HVT architecture in Sec.~\ref{sec:hvt} and talk about our method in detail in Sec.~\ref{sec:ourmethod}.
\vspace{-4mm}
\subsection{HVT Architecture}
\label{sec:hvt}
\vspace{-7mm}
\begin{figure}
\centering
\includegraphics[width=\linewidth, trim={1.5cm 3.0cm 8.2cm 0.8cm},clip]{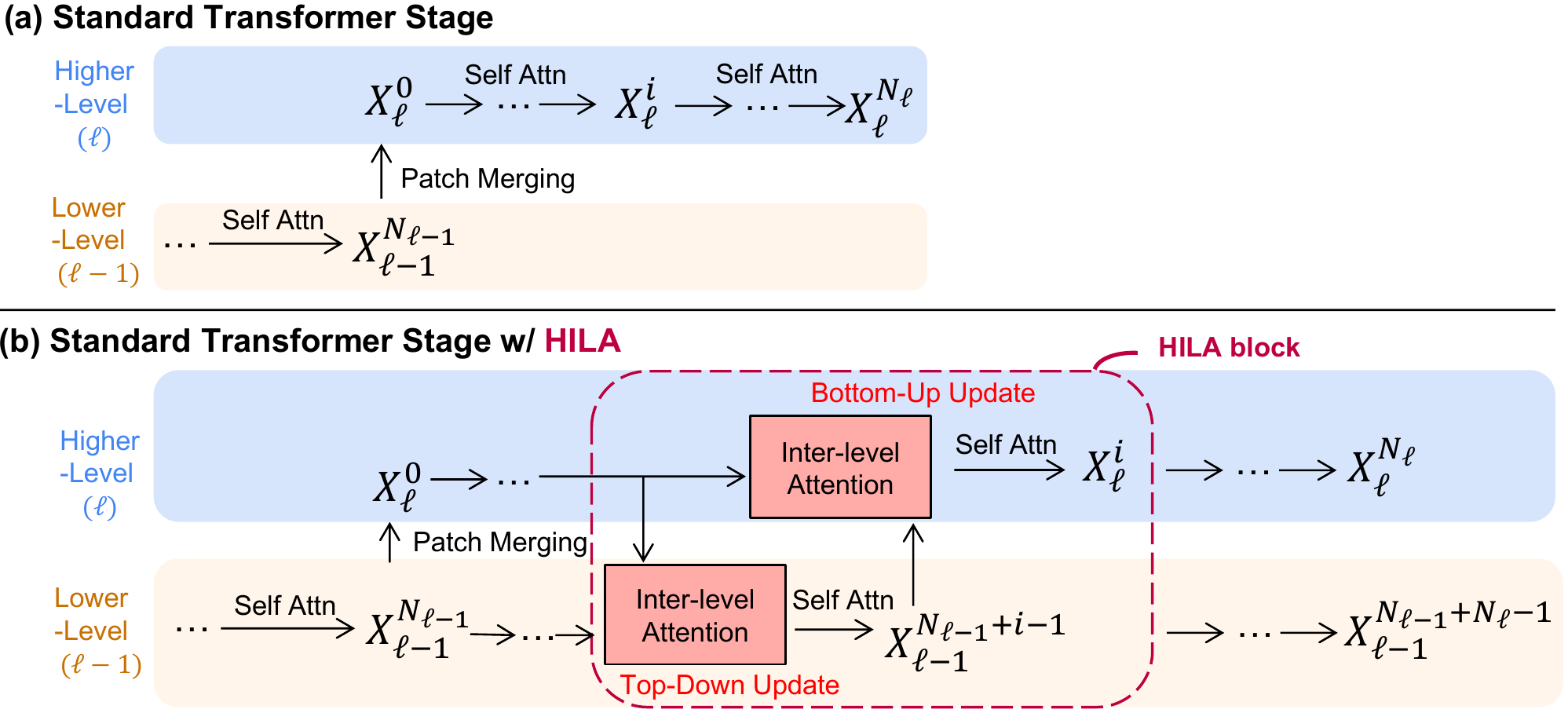} 
\vspace{-7mm}
\caption{\footnotesize \textbf{Our  \method{} Block.} (a) shows the standard hierarchical vision transformer block. (b) We wrap the transformer block with \method{}'s Bottom-Up and Top-Down Update. }
\label{fig:method_diagram}
\vspace{-7mm}
\end{figure}

In this section, we summarize the general HVT architecture for semantic segmentation and detail each HVT stage's components in general terms. The specific implementation may differ depending on the method.

Figure \ref{fig:method_diagram}(a) shows a stage of the standard HVT architectures.
Given an input image $I \in \mathbb{R}^{H\times W\times 3}$, where $H$ and $W$ denote the height and width of the image, HVT architectures sequentially process the image through multiple stages. In the semantic segmentation case, this is typically done in four features stages  with resolution scales at each stage of $\{\frac{1}{4},\frac{1}{8},\frac{1}{16},\frac{1}{32}\}$, respectively. Within one stage, the feature maps are also processed sequentially with Transformer blocks. We denote the features presented within a stage of the HVT model as $X_{\ell}^{i} \in \mathbb{R}^{H_{\ell}\times W_{\ell}\times d_{\ell}}$, where $\ell \in\{1, 2, 3, 4\}$ represents the model stage,  $H_{\ell}, W_\ell, d_\ell$ are the stage-specific height, width, and channel dimensions respectively, 
and $i\in \{1,...,N_\ell\}$ denotes the index of the intermediate feature representation after each individual block within stage $\ell$. Here $N_\ell$ denotes the total number of Transformer blocks in stage $\ell$. 
Each stage starts with features $X_{\ell}^{0}$ and output features $X_{\ell}^{N_\ell}$ after $N_\ell$ total blocks. 

The majority of HVT architectures \cite{hvt:chen2021regionvit,hvt:chen2021dpt,hvt:chu2021twins,hvt:dong2021cswin,hvt:fan2021multiscale,hvt:li2021localtoglobal,hvt:liu2021Swin,hvt:wang2021pyramid,hvt:wang2021pvtv2,hvt:scale:wang2021crossformer,hvt::wu2021cvt,hvt:wu2021pale,hvt:xie2021segformer,hvt:yang2021focal} follow the same design for each stage: 1) Propagate the feature from lower to higher stages through Patch Merging, and 2) Inside each stage, the feature map is refined through a sequence of Transformer Blocks. 

\textit{\textbf{1) Patch Merging.}} Given an input feature map, $X_{\ell-1}^{N_{\ell-1}}$, which is generated from the previous stage $\ell-1$ 
(or the image $I$ if $l=1$), 
the Patch Merging step downsamples 
the input feature 
and projects the feature dimensions for the next stage $\ell$. This is represented as $\mathbb{R}^{H_{\ell-1} \times W_{\ell-1} \times d_{\ell-1}} \rightarrow \mathbb{R}^{H_{\ell} \times W_{\ell} \times d_{\ell}}$. Downsampling is typically implemented using a convolution layer.
Several HVT architectures \cite{hvt:liu2021Swin,hvt:wang2021pyramid,hvt:yang2021focal,hvt:chu2021twins} create non-overlapping patches through equal kernel size and stride, while others use a variation where patches overlap (convolutional downsampling) \cite{hvt:wang2021pvtv2,hvt:xie2021segformer,hvt::wu2021cvt}. The final output is $X_{\ell}^{0}$, the starting feature of the stage $\ell$.

\textit{\textbf{2) Transformer Block.}} The Transformer Block consists of several individual sub-layers. Each sub-layer is generally implemented as a form of self-attention or fully connected layer, with the design depending on the specific HVT architecture. 

\vspace{-3mm}
\subsection{\methodfull{}}
\label{sec:ourmethod}
\vspace{-3mm}

\method{} wraps each Transformer Block in one stage with connections between current and previous stages.  Our main intuition is that higher and lower stage features stand to mutually improve each other. Lower-stage features have higher resolution and preserve more boundary information and details of the image, but are noisier and lack global semantics due to less capacity and smaller receptive fields. We thus design the Top-Down Update to pass higher-stage information downwards to improve lower-stage features, resulting in better semantic boundaries. On the other side, higher-stage features have noisy initialization due to Patch Merging, which pools all initial lower-stage features in a local area.
We thus design Bottom-Up Update to selectively propagate updated lower-stage features upwards to improve the semantic clarity of higher-stage features.
We introduce these two main updates in Section~\ref{sec:bottom-up} \&~\ref{sec:top-dowm}.

\begin{figure}[t!]
\centering
\vspace{-3mm}
\includegraphics[width=0.7\linewidth, height=3.5cm, trim={3.0cm 3.1cm 7.6cm  1.9cm},clip]{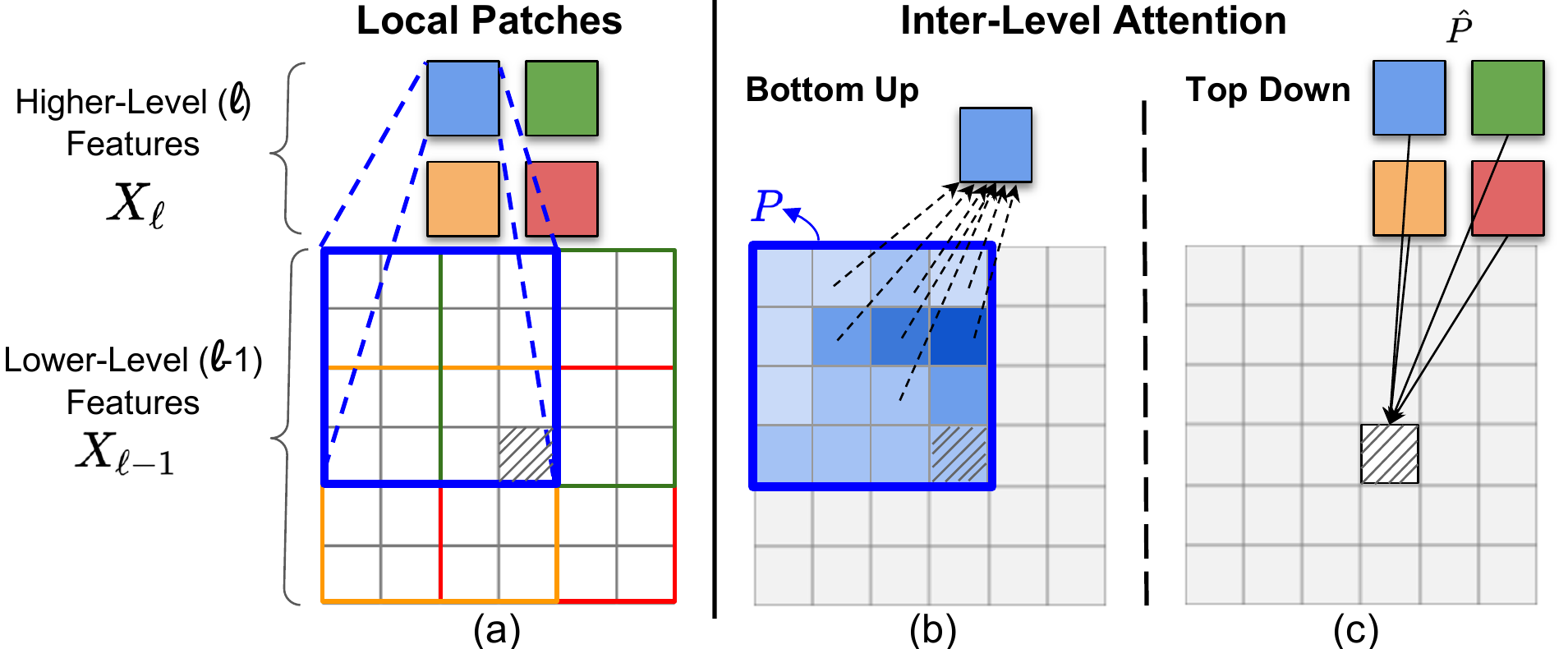} 
\vspace{-4mm}
\caption{\footnotesize \textbf{Top-Down and Bottom-Up Inter-Level Attention.} (a) Each higher-level feature corresponds to a local patch of lower-level features. (b) In our Bottom-Up Update, inter-level attention updates each higher-level feature using lower-level features in its local patch area $P$. Color indicates the attention weights of the high-level patch w.r.t each local patch in $P$. Darker color indicates the local patch have higher weights and will contribute more in the Bottom-Up Update.  (c) In our Top-Down Update, inter-level attention updates each lower-level feature using higher-level features whose patches they belong in. The hatched patch is covered by 4 higher level features $\hat{P}$.} 
\label{fig:patchgraph}
\vspace{-6mm}
\end{figure}

\newcommand{\hi}[1]{\ell}
\newcommand{\lo}[1]{\ell-1}

Figure \ref{fig:method_diagram}(b) illustrates the additions of \method{} to the standard HVT stage. 
Given a base HVT architecture, we apply \method{} to different stages.
For example, \methodthree{} means that we apply our method to stage 2, 3, and 4 of the base model. 
In particular, if \method{} is applied to stage $\ell$, we term the current stage $\ell$ as our higher-level and the previous stage $\ell-1$ as our lower-level. 
Following this, at iteration $i \in \{ 0,...,N_{\hi{}} \}$, our updated higher-level (current stage) feature is $X^{i}_{\hi{}}$, and our updated lower-level (previous stage) feature is $X^{N_{\lo{}}+i}_{\lo{}}$.  Because each stage is processed sequentially, the previous stage features $X_{\lo{}}$ has already been propagated through $N_{\lo{}}$ blocks in stage $\lo{}$, and then $i$ Top-Down Update blocks, resulting in a count of $N_{\lo{}}+i$. The final output of the stage $\ell-1$ will be $X_{\ell-1}^{N_{\ell-1} + N_{\ell}-1}$, as shown in \uncertain{Figure \ref{fig:method_diagram}(b)}. 

\vspace{-2mm}
\subsection{Bottom-Up Update}
\label{sec:bottom-up}
\vspace{-2mm}
Our Bottom-Up Update uses lower-level features to improve the higher-level features within a local patch area. This update consists of Bottom-Up Inter-Level Attention, where we propagate lower-level features to our higher-level features, followed by a Self-attention Layer, where we refine our higher-level features.

\paragraph{Bottom-Up Inter-Level Attention.} For one specific location within the higher-level feature map, $X^{i}_{\hi{}, \{h, w\}}$, where $\{h,w\}$ denotes the 2D coordinates in the feature map, we first determine a local patch area $P_{hw}$ within the lower-level features map $X^{N_{\lo{}}+i}_{\lo{}}$, and denote it as $X^{N_{\lo{}}+i}_{\lo{}, P_{hw}}$. This local area represents possible locations that the higher-level feature could look at in the lower-level feature level. We choose to use a $4\times4$ patch centered at location $\{h, w\}$ for $P_{hw}$, resulting in each higher-level feature corresponding to 16 lower-level features. 
We illustrate this correspondence in \uncertain{Figure~\ref{fig:patchgraph}(a)}.  

Our lower-level features within the local patch compete for assignment weights for the specific higher-level feature, as shown in \uncertain{Fig.~\ref{fig:patchgraph}(b)}. The motivation behind this is that lower-level features that are semantically aligned to the higher-level features will have more influence when propagated upwards and thus have larger attention weights. 
This allows us to improve the semantic clarity of the higher-level features in each update as these features will see updates favoring the lower-level ``components" that agree with them, in contrast to the noisy initialization from Patch Merging.  

Our Inter-Level Attention step uses a standard form of dot-product attention as described in~\cite{vaswani2017attention}. The inputs to our attention are our higher-level feature $X^{i}_{\hi{}, \{h, w\}}\in \mathbb{R}^{1\times d_{\hi{}}}$ and our lower-level feature patch $X^{N_{\lo{}}+i}_{\lo{}, P_{hw}}\in \mathbb{R}^{16\times d_{\lo{}}}$. To simplify our notation, we drop the 2D coordinate designations $\{h, w\}$ in the following equations. We project them into the query $Q = q(X^{i}_{\hi{}})$,  key $K = k(X^{N_{\lo{}}+i}_{\lo{}, P})$ and value $V = v(X^{N_{\lo{}}+i}_{\lo{}, P_{}})$, where $q, k, v$ are full-connected layers shared across different pixels. We then compute the attention using the following operation:
\vspace{-3mm}
\begin{equation}
\label{eqn:selfattention}
\vspace{-1mm}
\text{attn}(X^{i}_{\hi{}}, X^{N_{\lo{}}+i}_{\lo{}, P_{}}) = \text{Softmax}(\frac{QK^{\top}}{\sqrt{d_\ell}} + B)V, 
\vspace{-3mm}
\end{equation}
where we follow Swin Transformer~\cite{hvt:liu2021Swin} and add the relative positional encoding $B$. Our corresponding Bottom-Up Inter-Level Attention update is then: 
\vspace{-2mm}
\begin{eqnarray}
X^{i\prime}_{\hi{}}&=& X^{i}_{\hi{}} +  \text{attn}(X^{i}_{\hi{}}, X^{N_{\lo{}}+i}_{\lo{}, P_{}}),\\ 
X^{i\prime\prime}_{\hi{}} &=&\alpha X^{i\prime}_{\hi{}} + \beta \text{FFN}(X^{i\prime}_{\hi{}}),
\end{eqnarray} 
where $i'$ and $i''$ denote the intermediate outputs of our Bottom-Up Update, and $\alpha$ and $\beta$ are hyper-parameters controlling how much information to carry over from the previous iteration. We follow standard practice~\cite{hvt:xie2021segformer} and refine our features using a feed-forward network $\text{FFN}$ after the attention step. We provide further details in the \uncertain{Appendix}.



\vspace{-1mm}
\paragraph{Self-Attention Layer.} After updating higher-level features using Inter-Level Attention, we then propagate this information between features within the same level. We accomplish this by directly utilizing the self-attention layer from the base HVT model's Transformer Blocks: 
\begin{equation}
\vspace{-2mm}
X^{i+1}_{\hi{}} =  \text{SelfAttn}(X^{i\prime\prime}_{\hi{}}),
\end{equation}
where SelfAttn represents the self-attention layer from the base HVT model~\cite{hvt:liu2021Swin,hvt:xie2021segformer}. This layer processes features exactly the same as a typical Transformer Block would in the base HVT model, and as a result, no changes are required to the base HVT model. This makes \method{} flexible and easy to add to the majority of HVT architectures. 

\vspace{-3mm}
\subsection{Top-Down Update}
\label{sec:top-dowm}
\vspace{-1mm}
The Top-Down Update mirrors the Bottom-Up Update and propagates the higher-level features to update the lower-level features.
\vspace{-3mm}
\paragraph{Top-Down Inter-Level Attention Layer.} 
In reverse 
of the Bottom-Up Update, the Top-Down Update uses higher-level features to update the lower-level features. Each lower-level features  $X^{N_{\lo{}}+i}_{\lo{}, \{h,w\}}\in \mathbb{R}^{1\times d_{\lo{}}}$ is covered by the local patch areas of up to 4 higher-level features  $X^{i}_{\hi{}, \hat{P}_{hw}} \in \mathbb{R}^{n\times d_{\ell}}$, where $\hat{P}_{hw}$ denotes locations of higher-level features and $n \in\{
1,2,3,4\}$. The majority of lower-level features will be covered by 4 higher-level features (\textit{i.e.}, $n=4$), with the image boundaries having as few as 1 feature (\textit{i.e.}, $n=1$). 

Our Top-Down Inter-Level Attention is illustrated in \uncertain{Figure \ref{fig:patchgraph}(c)}. We treat each lower-level feature as semantically ``belonging" to one of the higher-level features whose local area it is under. Higher-level features compete against each other in the attention update, and semantically similar higher-level features will align the lower-level features to it. As a result, the lower-level feature can use this higher-level information to improve and correct its own representation. Similar to the above, we drop our coordinate notation for better clarity in our equations. Our Top-Down Inter-Level Attention update is defined as:\vspace{-2mm}
\begin{eqnarray}
\vspace{-2mm}
X^{N_{\lo{}}+i\prime}_{\lo{}} &=& X^{N_{\lo{}}+i}_{\lo{}} +  \text{attn}(X^{N_{\lo{}}+i}_{\lo{}}, X^{i}_{\hi{}, \hat{P}_{}})\\
X^{N_{\lo{}}+i\prime\prime}_{\lo{}} &=& \alpha X^{N_{\lo{}}+i\prime}_{\lo{}} + \beta \text{FFN}(X^{N_{\lo{}}+i\prime}_{\lo{}})
\vspace{-3mm}
\end{eqnarray}
\vspace{-11mm}
\paragraph{Self-Attention Layer.} Once again, after updating our lower-level features with our Inter-Level Attention, we would like to propagate this information between features of the same level. In the Top-Down Update, there is no pre-existing Transformer Block to integrate. As a result, we initialize a new block for this step, reusing the design of the Transformer Block from the baseline architecture. Our self-attention layer update is then defined as: \vspace{-2mm}
\begin{equation}
\vspace{-3mm}
X^{N_{\lo{}}+i+1}_{\lo{}} =  \text{SelfAttn}(X^{N_{\lo{}}+i\prime\prime}_{\lo{}})
\vspace{-3mm}
\end{equation}
\vspace{-3mm}
\subsection{Overall \method{} Block}
\vspace{-3mm}
We illustrate the \method{} block in \uncertain{Figure \ref{fig:method_diagram}(b)}. The \method{} block consists of a Top-Down Update followed by a Bottom-Up Update, with the exception being a stage's first \method{} block, where we skip the Top-Down Update. This is because the first \method{} block occurs immediately after Patch Merging and the newly formed higher-level features are too noisy to pass down meaningful information to the lower-level features.  

Given the initial higher-level features $X^{0}_{\hi{}}$ and lower-level features $X^{N_{\lo{}}}_{\lo{}}$, our first \method{} Block ($i=1$) skips the Top-Down Update and only applies our Bottom-Up Update. This improves the initial higher-level features which have been initialized with the patch embedding, and results in updated higher-level features $X^{1}_{\hi{}}$. 
Subsequent \method{} blocks ($i > 1$) then follow up with both Top-Down and Bottom-Up Updates. In the Top-Down Update, the refined higher-level features are propagated into the lower-level features, resulting in updated lower-level features $X^{N_{\lo{}}+i}_{\lo{}}$ with improved semantic meaning. In the Bottom-Up Update, the updated lower-level features will be propagated to the higher-level features, resulting in higher-level features $X^{i}_{\hi{}}$ with further improved semantic clarity. This iterative cycle then repeats in the remaining \method{} blocks. Our final outputs for the higher-level $\ell$ and lower-level $\ell-1$ are $X^{N_\hi{}}_{\hi{}}$ and $X^{N_{\lo{}}+N_\hi{}-1}_{\lo{}}$, respectively. 

We share the weights of the extra component we added for the \method{} block within the same stage, \textit{i.e.}, in Figure~\ref{fig:method_diagram}(b), the weights of the red boxes are shared across different $i$. 
Intuitively, across all iterations, the \method{} block updates features between different levels through attention and refines assignment weights in an consistent manner. Another advantage of weight-sharing is that the number of parameters does not increase as the number of iterations increases. 
While \method{} blocks are independent of each other between different stages, we show in Section~\ref{sec:attentionvisualization} that the attention weights in \method{} across different stages can form a meaningful hierarchy of the whole object. We provide a complexity analysis in the Appendix.
\vspace{-3mm}
\subsection{Difference from Past Work}
\label{differencepastwork}
\vspace{-1mm}
\method{} aims to extend existing backbones by propagating information between levels – to resolve ambiguities
in the lower-level features and improve localization for the
higher-level features. Our design achieves this through
- 1) \textbf{explicitly} forcing selective information to be passed to
resolve ambiguities and creating hierarchies among objects
using an attention mechanism, and 2) \textbf{iteratively updating}
bi-directional information \textbf{multiple times} to allow the hierarchies to emerge and improve from the iterations instead of directly outputting the hierarchy in one step. The combination of these two unique designs helps HVTs with \method{} learn better features for disambiguating different objects. Prior works, such as HRNet~\cite{sun2019hrnet} and a parallel work HRViT~\cite{gu2022hrvit}, use convolution inter-level updates in a residual manner. This results in difficulty resolving these ambiguities as there is no explicit mechanism to force this behaviour in their CNN fusion module
- only that information is passed between levels. In
the case of \method{}, when passing inter-level features, dot-product attention forces the weights of each information
passed to sum to one, explicitly forcing features to compete
with each other to pass information to other levels based on
alignment in similarity. In addition, this alignment can be
iteratively updated and improved across multiple iterations
with our design of \method{}. This is in contrast to HRNet and HRViT, which only perform fusion once at the end
of each stage.

\vspace{-4mm}
\section{Experiments}
\vspace{-3mm}
\subsection{Experimental Settings}
\vspace{-2mm}
\paragraph{Datasets:} We evaluate our method using two publicly available datasets: Cityscapes \cite{cordts2016cityscapes} and ADE20K~\cite{ade20k}. Cityscapes is a driving dataset for semantic segmentation consisting of images from 27 cities in/near Germany. We use the fine-annotated high resolution images with 19 categories across a set of 2975 training, 500 validation, and 1525 test images. ADE20K contains generic scenes and is segmented into 150 semantic categories. The dataset consists of 25K images, of which 20K is for training, 2K for validation and 3K for testing.

\vspace{-2mm}
\paragraph{Architecture:}
Our \method{} module is compatible with the majority of HVT architectures. Our results build upon two architectures: SegFormer \cite{hvt:xie2021segformer} which currently holds the state-of-the-art in semantic segmentation and Swin-Transformer \cite{hvt:liu2021Swin}, a well-known architecture that serves as a strong baseline on several vision tasks. We evaluate in two settings: \methodone{} and \methodthree{}. We use \methodone{} for smaller baselines, and expand to our more complex hierarchical \methodthree{} method as we scale up our compute. e. 

\vspace{-2mm}
\paragraph{Training Settings:} We build upon SegFormer and Swin Transformer's implementation within the mmsegmentation repository \cite{mmseg2020}. For training, we follow the exact same procedure as the backbone model we augment.  Following Segformer~\cite{hvt:xie2021segformer}, our data is augmented with random resize with a ratio of 0.5-2.0, random horizontal flipping, and random cropping to either 512x512 for ADE20K or 1024x1024 for Cityscapes. We use a batch size of 16 for ADE20K and 8 for Cityscapes, and train our models in 4 Nvidia RTX6000 GPUs. We train all our models using the AdamW optimizer for an effective 160K iterations on both datasets, with an initial learning rate of 0.00006 and a ``poly" LR schedule with a default factor of 1.0. We do not use OHEM, auxiliary losses, or class balance loss. We detail model specific hyper-parameters in the Appendix. 

\vspace{-2mm}
\paragraph{Pretraining:} We experiment with two versions of our models with one having \method{} pretrained and one without. To pretrain \method{}, we train the backbone model with \method{} together from scratch on Imagenet1K following the exact same procedure as the backbone model we augment. In the case that \method{} is not pretrained, we reuse pre-training weights provided from the official repository of the backbone model we use. \method{} would then be initialized randomly, with the exception being the Top-Down Update's self-attention layer, which copies the pre-trained weights of the backbone model's final self-attention layer in the previous stage. We note that \method{} directly changes and improves the backbone HVT itself. As a result, similar to many backbone network architectures used in segmentation methods, \method{} requires pretraining for the best performance . Due to a lack of computational resources, we were unable to pretrain larger models. We provide results with and without pretraining to show that better results can be achieved given more computational resources for pretraining in larger models. 
We provide results with and without pretraining to show that better results can be achieved given more computational resources for pretraining in larger models. 
\setlength{\tabcolsep}{4pt}
\begin{table}[t!]
\begin{center}
\vspace{-8mm}
\caption{\footnotesize \textbf{mIoU/F-score on ADE20K \& Cityscapes.} We apply \method{} to both Swin-T~\cite{hvt:liu2021Swin} and Segformer~\cite{hvt:xie2021segformer}. Our models achieve significant improvements over baselines in both mIoU and F-score on two datasets. With more inter-layer interactions, \methodthree{} achieves best performance. 
SS/MS refers to single scale/multi-scale test. 
}
\begin{adjustbox}{width=\textwidth}
\label{table:mainresults}
\begin{tabular}{|c|c|c|c|ccc|ccc|}
\hline
\multirow{2}{*}{\textbf{Method}} &
  \multirow{2}{*}{\textbf{Encoder}} &
  \multirow{2}{*}{\textbf{\begin{tabular}[c]{@{}c@{}}\method\\ Pretrained\end{tabular}}} &
  \multirow{2}{*}{\textbf{Params (M)}} &
  \multicolumn{3}{c|}{\textbf{ADE20K}} &
  \multicolumn{3}{c|}{\textbf{Cityscapes}} \\ \cline{5-10} 
 &
   &
   &
   &
  \multicolumn{1}{c|}{FLOPS} &
  \multicolumn{1}{c|}{\begin{tabular}[c]{@{}c@{}}mIOU \\ (SS/MS)\end{tabular}} &
  \begin{tabular}[c]{@{}c@{}}F-Score \\ 3px (SS)\end{tabular} &
  \multicolumn{1}{c|}{FLOPS} &
  \multicolumn{1}{c|}{\begin{tabular}[c]{@{}c@{}}mIOU \\ (SS/MS)\end{tabular}} &
  \begin{tabular}[c]{@{}c@{}}F-Score \\ 3px (SS)\end{tabular} \\ \hline
FCN \cite{long2015fcn} &
  ResNet-101 \cite{he2016resnet} &
  N/A &
  68.5 &
  \multicolumn{1}{c|}{275.4} &
  \multicolumn{1}{c|}{39.9 / 41.4} &
  ------ &
  \multicolumn{1}{c|}{2203.3} &
  \multicolumn{1}{c|}{75.5 / 76.6} &
  60.0 \\
PSPNet \cite{zhao2017pspnet} &
  ResNet-101 &
  N/A &
  68.0 &
  \multicolumn{1}{c|}{256.2} &
  \multicolumn{1}{c|}{43.6 / 44.4} &
  ------ &
  \multicolumn{1}{c|}{2048.9} &
  \multicolumn{1}{c|}{79.8 / 81.0} &
  73.6 \\
UperNet \cite{xiao2018upernet} &
  ResNet-101 &
  N/A &
  85.4 &
  \multicolumn{1}{c|}{256.3} &
  \multicolumn{1}{c|}{43.8 / 44.9} &
  ------ &
  \multicolumn{1}{c|}{2049.8} &
  \multicolumn{1}{c|}{79.4 / 80.5} &
  74.2 \\
CCNet \cite{huang2019ccnet} &
  ResNet-101 &
  N/A &
  68.9 &
  \multicolumn{1}{c|}{278.4} &
  \multicolumn{1}{c|}{43.7 / 45.0} &
  ------ &
  \multicolumn{1}{c|}{2224.8} &
  \multicolumn{1}{c|}{79.5 / 80.7} &
  73.7 \\
DANet \cite{fu2019danet} &
  ResNet-101 &
  N/A &
  68.8 &
  \multicolumn{1}{c|}{276.8} &
  \multicolumn{1}{c|}{43.6 / 45.1} &
  ------ &
  \multicolumn{1}{c|}{------} &
  \multicolumn{1}{c|}{80.5 / 82.0} &
  74.6 \\
DeeplabV3+ \cite{chen2018deeplabv3+} &
  ResNet-101 &
  N/A &
  62.7 &
  \multicolumn{1}{c|}{255.1} &
  \multicolumn{1}{c|}{------ / 44.1} &
  ------ &
  \multicolumn{1}{c|}{2032.3} &
  \multicolumn{1}{c|}{------ / 80.9} &
  ------ \\
OCRNet \cite{yuan2020ocrnet} &
  HRNet-W48 \cite{sun2019hrnet} &
  N/A &
  70.5 &
  \multicolumn{1}{c|}{164.8} &
  \multicolumn{1}{c|}{------ / 45.6} &
  ------ &
  \multicolumn{1}{c|}{1296.8} &
  \multicolumn{1}{c|}{------ / 81.1} &
  ------ \\
FCN \cite{long2015fcn} &
  D3Net-L \cite{takahashi2021d3net} &
  N/A &
  38.7 &
  \multicolumn{1}{c|}{------} &
  \multicolumn{1}{c|}{------} &
  ------ &
  \multicolumn{1}{c|}{------} &
  \multicolumn{1}{c|}{80.6 / ------} &
  ------ \\
GSCNN \cite{takikawa2019gatedscnn} &
  WideResNet38 \cite{zagoruyko2016wideresnet} &
  N/A &
  ------ &
  \multicolumn{1}{c|}{------} &
  \multicolumn{1}{c|}{------} &
  ------ &
  \multicolumn{1}{c|}{------} &
  \multicolumn{1}{c|}{80.8 / ------} &
  73.6 \\ \hline
Upernet \cite{xiao2018upernet} &
  Swin-T \cite{hvt:liu2021Swin} &
  N/A &
  59.8 &
  \multicolumn{1}{c|}{235.7} &
  \multicolumn{1}{c|}{44.4 / 45.8} &
  73.2 &
  \multicolumn{1}{c|}{} &
  \multicolumn{1}{c|}{------} &
  ------ \\ 
(Ours) &
  + \methodthree{} &
   &
  68.5 &
  \multicolumn{1}{c|}{266.02} &
  \multicolumn{1}{c|}{\textbf{44.9 / 46.1}} &
  \textbf{75.1} &
  \multicolumn{1}{c|}{------} &
  \multicolumn{1}{c|}{------} &
  ------ \\ \hline
SegFormer \cite{hvt:xie2021segformer} &
  MiT-B0 &
  N/A &
  3.7 &
  \multicolumn{1}{c|}{8.4} &
  \multicolumn{1}{c|}{37.4 / 38.0} &
  67.2 &
  \multicolumn{1}{c|}{121.2} &
  \multicolumn{1}{c|}{76.2 / 78.1} &
  68.6 \\ 
(Ours) &
  + \methodone{} &
   &
  4.2 &
  \multicolumn{1}{c|}{9.8} &
  \multicolumn{1}{c|}{38.3 / 38.8} &
  69.3 &
  \multicolumn{1}{c|}{139.4} &
  \multicolumn{1}{c|}{77.2 / \textbf{79.1}} &
  69.0 \\
 &
  + \methodone{} &
  \checkmark &
  4.2 &
  \multicolumn{1}{c|}{9.8} &
  \multicolumn{1}{c|}{\textbf{40.3 / 40.9}} &
  \textbf{69.4} &
  \multicolumn{1}{c|}{139.4} &
  \multicolumn{1}{c|}{ \textbf{77.3} / 78.7} &
  \textbf{70.3} \\ \hline
SegFormer &
  MiT-B1 &
  N/A &
  13.7 &
  \multicolumn{1}{c|}{15.9} &
  \multicolumn{1}{c|}{42.2 / 43.1} &
  70.6 &
  \multicolumn{1}{c|}{239.5} &
  \multicolumn{1}{c|}{78.5 / 80.0} &
  71.4 \\ 
(Ours) &
  + \methodone{} &
   &
  16.3 &
  \multicolumn{1}{c|}{20.9} &
  \multicolumn{1}{c|}{42.9 / 43.7} &
  72.0 &
  \multicolumn{1}{c|}{294.5} &
  \multicolumn{1}{c|}{78.9 / 80.2} &
  72.6 \\
 &
  + \methodone{} &
  \checkmark &
  16.3 &
  \multicolumn{1}{c|}{20.9} &
  \multicolumn{1}{c|}{44.0 / 45.0} &
  72.3 &
  \multicolumn{1}{c|}{294.5} &
  \multicolumn{1}{c|}{80.2 / 81.7} &
  73.4 \\
 &
  + \methodthree{} &
  &
  21.6 &
  \multicolumn{1}{c|}{31.4} &
  \multicolumn{1}{c|}{43.5 / 44.2} &
  73.5 &
  \multicolumn{1}{c|}{417.8} &
  \multicolumn{1}{c|}{79.9 / 81.3} &
  73.6 \\
 &
  + \methodthree{} &
  \checkmark &
  21.6 &
  \multicolumn{1}{c|}{31.4} &
  \multicolumn{1}{c|}{\textbf{45.4 / 46.2}} &
  \textbf{74.1} &
  \multicolumn{1}{c|}{417.8} &
  \multicolumn{1}{c|}{\textbf{80.9 / 82.4}} &
  \textbf{74.8} \\ \hline
SegFormer &
  MiT-B2 &
  N/A &
  27.4 &
  \multicolumn{1}{c|}{62.4} &
  \multicolumn{1}{c|}{46.5 / 47.5} &
  74.0 &
  \multicolumn{1}{c|}{704.2} &
  \multicolumn{1}{c|}{81.0 / 82.2} &
  74.2 \\ 
(Ours) &
  + \methodthree{} &
   &
  30.8 &
  \multicolumn{1}{c|}{76.5} &
  \multicolumn{1}{c|}{46.0 / 46.8} &
  \textbf{74.5} &
  \multicolumn{1}{c|}{867.4} &
  \multicolumn{1}{c|}{\textbf{81.5 / 82.6}} &
  \textbf{74.9} \\ \hline
\end{tabular}
\end{adjustbox}
\end{center}
\vspace{-9mm}
\end{table}
\setlength{\tabcolsep}{1.4pt}

\vspace{-7mm}
\paragraph{Evaluation Settings:} We evaluate the performance using two metrics: mean Intersection over Union (mIoU) and boundary F-score~\cite{davisPerazzi2016}, which focuses more on the boundary quality than mIoU. For mIoU, we follow Segformer~\cite{hvt:xie2021segformer} and evaluate under both single-scale (SS) and multi-scale (MS) inference. For F-score, we follow G-SCNN~\cite{takikawa2019gatedscnn} and compute F-score using the strictest threshold of 0.00088 which corresponds to a 3 pixel threshold for Cityscapes. For F-Score in ADE20K, we calculate the 3px boundary precision/recall for all classes at once instead of separately. This allows the metric to capture the overall boundary quality of the segmentation, while avoiding variance issues caused by a large number of semantic categories in ADE20K. For image inference in ADE20K, we follow Segformer~\cite{hvt:xie2021segformer} and rescale the shortest side of the image to the training cropping size and maintain the aspect ratio of original image. Further details are provided in the Appendix. 


\begin{table}[t!]
\begin{center}
\vspace{-6mm}
\caption{\footnotesize \textbf{Real-time inference comparison on Cityscapes.} Adding \method{} to Segformer gives significant improvements. All results reported on 1 Nvidia RTX6000 GPU.} 
\begin{adjustbox}{width=0.8\textwidth}
\label{table:realtimeresults}
\begin{tabular}{|c|c|c|c|cccc|}
\hline
\multirow{2}{*}{\textbf{Method}} &
  \multirow{2}{*}{\textbf{Encoder}} &
  \multirow{2}{*}{\textbf{Resolution}} &
  \multirow{2}{*}{\textbf{\begin{tabular}[c]{@{}c@{}}Params \\ (M)\end{tabular}}} &
  \multicolumn{4}{c|}{\textbf{Cityscapes}} \\ \cline{5-8} 
 &
   &
   &
   &
  \multicolumn{1}{c|}{FLOPS} &
  \multicolumn{1}{c|}{FPS} &
  \multicolumn{1}{c|}{\begin{tabular}[c]{@{}c@{}}mIOU \\ (MS)\end{tabular}} &
  \begin{tabular}[c]{@{}c@{}}F-Score \\ 3px (SS)\end{tabular} \\ \hline
FCN \cite{long2015fcn}                        & MobileNetV2 \cite{sandler2018mobilenetv2}                 & 1024x2048 & 9.8  & \multicolumn{1}{c|}{317.1} & \multicolumn{1}{c|}{11.6}    & \multicolumn{1}{c|}{61.5} & 65.1    \\ \hline
PSPNet \cite{zhao2017pspnet}                     & MobileNetV2                  & 1024x2048 & 13.7 & \multicolumn{1}{c|}{139.4} & \multicolumn{1}{c|}{10.1}    & \multicolumn{1}{c|}{70.9} & 66.5    \\ \hline
DeepLabV3+ \cite{chen2018deeplabv3+}                 & MobileNetV2                  & 1024x2048 & 15.4 & \multicolumn{1}{c|}{239.5} & \multicolumn{1}{c|}{7.3}    & \multicolumn{1}{c|}{76.3} & 67.74    \\ \hline
\multirow{4}{*}{SegFormer} & \multirow{4}{*}{MiT-B0}      & 1024x2048 & 3.8  & \multicolumn{1}{c|}{125.5} & \multicolumn{1}{c|}{7.0}  & \multicolumn{1}{c|}{78.1} & 68.6 \\
                           &                              & 768x1536  & 3.8  & \multicolumn{1}{c|}{51.7}  & \multicolumn{1}{c|}{20.4}  & \multicolumn{1}{c|}{77.6} & 66.2 \\
                           &                              & 640x1280  & 3.8  & \multicolumn{1}{c|}{31.5}  & \multicolumn{1}{c|}{26.5} & \multicolumn{1}{c|}{75.3} & 65.0 \\
                           &                              & 512x1024  & 3.8  & \multicolumn{1}{c|}{17.7}  & \multicolumn{1}{c|}{41.5} & \multicolumn{1}{c|}{74.2} & 63.1 \\ \hline
\multirow{4}{*}{(Ours)}    & \multirow{4}{*}{+ \methodone{}} & 1024x2048 & 4.4  & \multicolumn{1}{c|}{139.4} & \multicolumn{1}{c|}{5.7}  & \multicolumn{1}{c|}{\textbf{78.7}} & \textbf{70.1} \\
                           &                              & 768x1536  & 4.4  & \multicolumn{1}{c|}{59.9}  & \multicolumn{1}{c|}{15.8}  & \multicolumn{1}{c|}{\textbf{78.4}} & \textbf{68.2} \\
                           &                              & 640x1280  & 4.4  & \multicolumn{1}{c|}{36.6}  & \multicolumn{1}{c|}{20.6}  & \multicolumn{1}{c|}{\textbf{77.0}} & \textbf{67.5} \\
                           &                              & 512x1024  & 4.4  & \multicolumn{1}{c|}{20.8}  & \multicolumn{1}{c|}{31.7} & \multicolumn{1}{c|}{\textbf{76.8}} & \textbf{65.5} \\
                           \hline
\end{tabular}
\end{adjustbox}
\end{center}
\vspace{-11mm}
\end{table}

\vspace{-6mm}
\subsection{Comparison to State-of-the-Art methods}
\vspace{-3mm}
We first compare our \method{} against state-of-the-art methods by adding \method{} to SegFormer architecture in both ADE20K and Cityscapes. We report quantitative results in Table~\ref{table:mainresults}, with qualitative examples in Figure~\ref{figure:qualitativeevaluation}. Adding our pretrained \method{} to the SegFormer baselines considerably increases performance in terms of both mIoU and F-score. Specifically, the SegFormer B1 + \methodthree{} model increases performance over baseline SegFormer B1 by 2.4 mIoU / 3.4 F-score. When comparing  SegFormer B1 + \methodthree{} to the larger model of SegFormer B2, we surpass it by 0.2 mIoU / 0.6 F-score while being significantly more efficient, with 21\% less parameters and 41\% less FLOPS. Similar improvements can also be observed by adding \methodone{} to smaller sized models, such as the SegFormer B0 model.

We then show a class-by-class comparison between the baseline B1 model with and without \methodthree{} in Figure~\ref{figure:perclasscomparison}, and report the exact numbers in the Appendix. We roughly divide the 19 classes in Cityscapes into three categories: Background, Vehicles, and Thin/Complex objects. When using \methodthree{}, objects in the Thin/Complex category have the largest improvement, especially in terms of F-score, demonstrating the effectiveness of our method in extracting fine details for the object boundaries. Qualitatively speaking, when comparing our results with SegFormer~\cite{hvt:xie2021segformer} in Figure~\ref{figure:qualitativeevaluation}, our results have more precise boundaries and less errors between different semantic labels,  particularly for thinner and complex objects such as poles, people, bicycles, etc. We attribute this increase to \method{}'s ability to improve the representations in lower-level features, allowing our augmented models to be particularly effective in localizing the boundaries of smaller and complex objects. 

\begin{figure}[t!]
\vspace{-2mm}
\centering
\includegraphics[width=\linewidth, trim=0.5cm 10.0cm 3.5cm 0.5cm, clip]{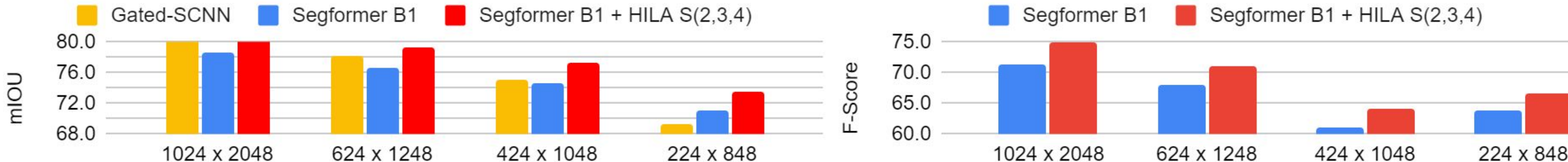} 
\vspace{-5mm}
\caption{\footnotesize \textbf{Distance-based Evaluation in Cityscapes.} We show mIoU/F-Score at various crop levels, with smaller crops weighting small and distant objects heavier.
}
\label{figure:distanceeval}
\vspace{-6mm}
\end{figure}

\vspace{-2.5mm}
\paragraph{Distance based evaluation:} Following Gated-SCNN \cite{takikawa2019gatedscnn}, we provide distance-based evaluations on Cityscapes. We evaluate crops of our outputs centered around the approximate vanishing point of our images, with smaller crops containing smaller and more distant objects. Figure \ref{figure:distanceeval} shows a crop by crop comparison between Gated-SCNN and the baseline B1 model with and without \methodthree{}. \methodthree{} outperforms the baseline at all crop distances, and maintain +2.4 mIoU (SS) and +2.8 F-score in the smallest crop of 224 by 848, which emphasizes distant objects the most. The F-score is a precision/recall based metric and increases in the smallest crop size due to decreased class diversity.

\vspace{-2.5mm}
\paragraph{Pretraining:} We note that, in Table~\ref{table:mainresults}, even without pretraining \method{} on ImageNet1K, we still achieve better performance comparing with baselines, which are fully pretrained on ImageNet1K. With pretraining \method{} on ImageNet1K, we obtain even higher improvements. In particular, for SegFormer B1, Adding \methodone{} without pretraining achieves +1.4 mIoU (MS) and +1.2 F-score improvements, and if adding pretraining, the improvements becomes +1.7 mIoU (MS) and +2.0 F-score. Due to the lack of computational resources, we were unable to pretrain larger models, such as SegFormer B2. With more computational resources to pretrain larger models, we expect better performance. 

\begin{figure*}[t!]
\centering
\vspace{-5mm}
\includegraphics[width=\linewidth, trim=0cm 0.0cm 0cm 0cm, clip]{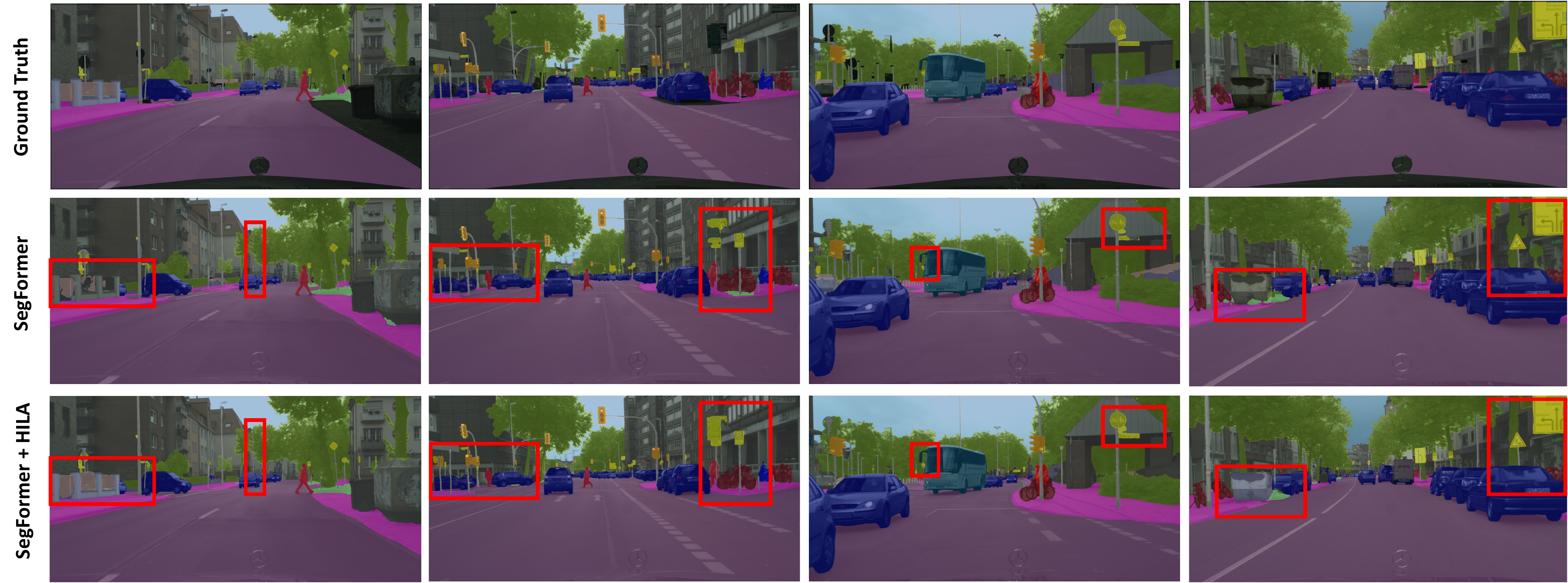} 
\vspace{-7mm}
\caption{\footnotesize \textbf{Qualitative comparison on Cityscapes.} We compare our method with SegFormer baseline and highlight the differences in red rectangles. Our results align better with object boundaries, especially for thin objects, such as poles. }
\label{figure:qualitativeevaluation}
\vspace{-4mm}
\end{figure*}

\vspace{-2.5mm}
\paragraph{Other Transformer Backbone:} We also add our method \methodthree{} into Swin Transformer~\cite{hvt:liu2021Swin} and train the network on ADE20K dataset. We achieve better performance (+0.5 mIoU/ +1.8 F-Score) even without pretraining \methodthree{}, supporting that our approach can be integrated into different HVT models.

\vspace{-2.5mm}
\paragraph{Real Time inference:}  In Table~\ref{table:realtimeresults}, we further report real-time inference results for our Segformer-B0 + \methodone{} model. When predicting on 768x1536 resolution images, our model yields 15.8 FPS and 78.4 mIoU, as compared to the baseline, which needs to run at 1024x2048 resolution to achieve similar performance (78.1 mIoU), but with much lower FPS (7.0).

\begin{figure}[t!]
\centering
\vspace{-0mm}
\includegraphics[width=0.95\linewidth, trim=1.5cm 4.1cm 0.5cm 1.2cm, clip]{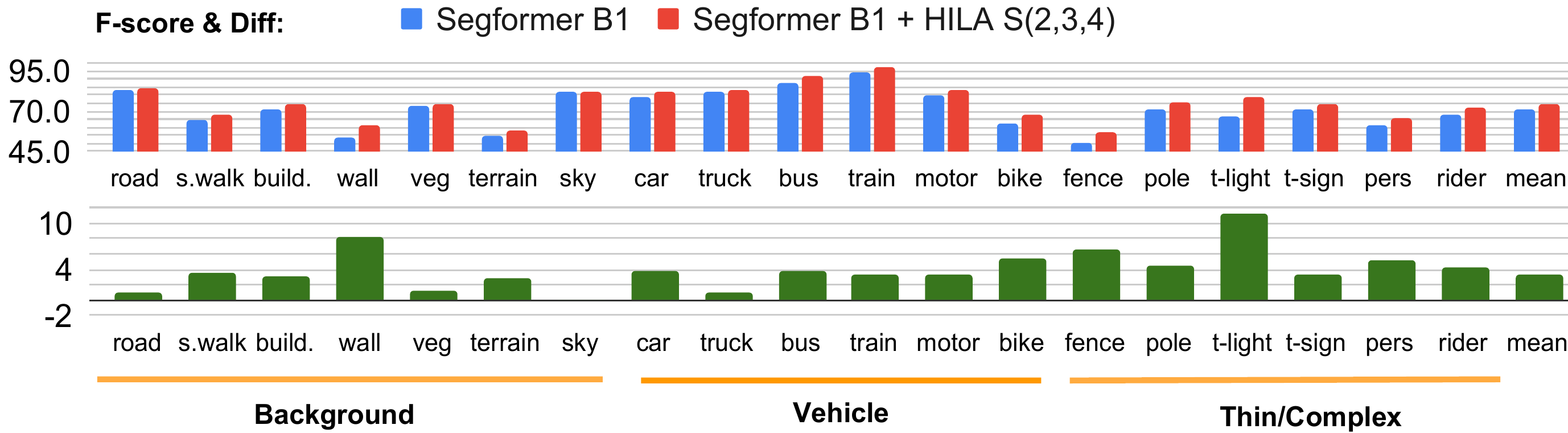} 
\vspace{-4mm}
\caption{\footnotesize \textbf{Class by class comparison in Cityscapes on F-score and F-score differences. } We show  improvements in adding \method{} S(2,3,4) to the Segformer B1 model.  
}
\label{figure:perclasscomparison}
\vspace{-2mm}
\end{figure}

\vspace{-4mm}
\subsection{Ablation Studies}
\vspace{-1mm}
We ablate our method by applying \method{} to different stages of SegFormer B1 and removing different components of \methodthree{} on Cityscapes. We use ImageNet1K pretrained model in this section.

\vspace{-2mm}
\paragraph{Ablating different stages:} 
We first ablate which stage we should apply \method{} to if we only apply it to one stage.  Results are reported in Table~\ref{table:ablation_stages}. We see the largest improvements in S(3). This is likely due to S(3) providing the best balance of feature capacity versus feature resolution in Cityscapes. One intuition supporting this idea is that HVT architectures generally scale the number of Stage 3 blocks as network depth increases. We note that S(2) creates a drop in performance as compared to the baseline. We reason that S(2) interacts between Stage 1 and 2, both of which are at shallow stages of a model and have low capacity. As a result, it is difficult to learn meaningful information to pass between these two levels. 
We further analyze the effects of adding our \method{} to different number of stages and report the results in Table \ref{table:ablation_stages}. 
We observe continual improvements when we add \method{} to more stages: from S(4) to S(3,4) and S(2,3,4). 
When \method{} is added to different levels, the final prediction can be improved in multiple levels of detail, resulting in a much better performance. 


\vspace{-2mm}
\paragraph{Ablating different components:} We now analyze the effects of each component in \method{}. We add our Bottom-Up and Top-Down Updates one at a time to the SegFormer B1 model, and  report the results in Table~\ref{table:ablation_effectofupdates}. We observe improvements to the baseline by adding our Bottom-Up Inter-Level Attention. In particular, we see +1.4 and +2.6  improvement in terms of mIoU and F-score, respectively, showing the effectiveness of our Bottom-Up Update module. When we further add the Top-Down Update on top of it, the improvement becomes more significant, demonstrating the effects of our method in jointly improving both higher-level and lower-level features. 

\begin{table}[]
\vspace{-8mm}
\caption{\footnotesize \textbf{Ablation studies of Our Method on Cityscapes.} We first apply \method{} to different stages and show the results on the left, and then ablate different components of our \method{} on the right. Both ablated on Segformer B1 + \methodthree{} model.}
\vspace{-4mm}
\begin{subtable}{.5\linewidth}
\caption{\footnotesize Apply \method{} to different stages}
\vspace{-4mm}
\begin{center}
\vspace{-1mm}
\begin{adjustbox}{width=\textwidth}
\label{table:ablation_stages}
\begin{tabular}{|c|c|c|c|c|c|}
\hline
\textbf{Name} &
  \textbf{Stage 2} &
  \textbf{Stage 3} &
  \textbf{Stage 4} &
  \textbf{\begin{tabular}[c]{@{}c@{}}mIOU \\ (SS)\end{tabular}} &
  \textbf{\begin{tabular}[c]{@{}c@{}}F-score \\ (SS)\end{tabular}} \\ \hline
N/A      &           &           &           & 78.5          & 71.4          \\
S(2)     & \cmark &           &           & 77.8          & 70.9          \\
S(3)     &           & \cmark &           & \textbf{80.2} & \textbf{73.4} \\
S(4)     &           &           & \cmark & 78.5          & 71.8          \\
S(3,4)   &           & \cmark & \cmark & 80.6          & 73.8          \\
S(2,3,4) & \cmark & \cmark & \cmark & \textbf{80.8} & \textbf{74.5} \\ \hline
\end{tabular}
\end{adjustbox}
\vspace{-7mm}
\end{center}
\end{subtable}
\begin{subtable}{.5\linewidth}
\caption{\footnotesize Ablate different components of \method{}}
\vspace{-4mm}
\begin{center}
\begin{adjustbox}{width=0.9\textwidth}
\label{table:ablation_effectofupdates}
\begin{tabular}{|c|c|c|c|}
\hline
\textbf{\begin{tabular}[c]{@{}c@{}}Bottom-Up \\ Update\end{tabular}} & \textbf{\begin{tabular}[c]{@{}c@{}} Top-Down \\ Update\end{tabular}} & \textbf{\begin{tabular}[c]{@{}c@{}} mIoU \\ (SS)\end{tabular}} & \textbf{\begin{tabular}[c]{@{}c@{}} F-score \\ (SS)\end{tabular}} \\ \hline

   \xmark       &         \xmark    & 78.5 & 71.4 \\
\cmark &     \xmark    & 79.9 & 73.0 \\
\cmark &  \cmark & \textbf{80.8} & \textbf{74.5} \\ \hline
\end{tabular}
\end{adjustbox}
\end{center}
\end{subtable}
\vspace{-4mm}
\end{table}

\vspace{-3mm}
\subsection{Hierarchical Visualization}
\label{sec:attentionvisualization}
\vspace{-2mm}
In the Top-Down Update, the attention weights represent the assignments of higher-level features into the lower-level features. If we progressively multiply the weight matrix from the top-most level to bottom-most, we can generate a hierarchy corresponding from the object whole to the object parts. Specifically, 
each Stage 4 feature, $X^{N_4}_{4,\{h, w\}}$, assigns itself to a local patch of lower-level Stage 3 features $X^{N_3+N_4-1}_{3, P_{}}$, and similarly, each Stage 3 feature also has the attention weights to assign itself to the Stage 2 features, and Stage 2 features to Stage 1. The magnitude of the attention weights represents the alignment between lower-level features and higher-level  features. We combine these attention weights in a hierarchical manner by multiplying them. For example, an attention weight of 0.3 between the Stage 4 feature and a specific Stage 3 feature would propagate downwards, multiplying with the attention weights between that specific Stage 3 feature and a Stage 2 feature (say 0.2), resulting in a final Stage 4 to Stage 2 weight of 0.06. This can propagate again from Stage 2 to Stage 1 and we then re-scale the final hierarchical attention to obtain the final attention weights from Stage 4 to Stage 1. We provide a detailed explanation with mathematical notation in the Appendix.

\begin{figure}[t!]
\vspace{-5mm}
\centering
\includegraphics[width=0.08\textwidth]{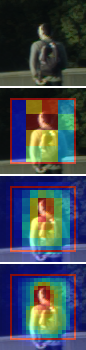}
\includegraphics[width=0.08\textwidth]{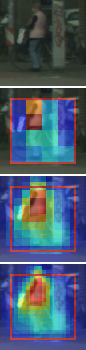}
\includegraphics[width=0.08\textwidth]{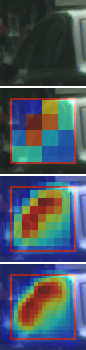}
\includegraphics[width=0.08\textwidth]{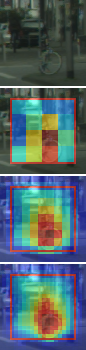}
\includegraphics[width=0.08\textwidth]{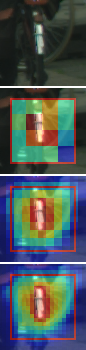}
\includegraphics[width=0.08\textwidth]{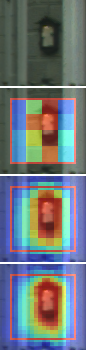}
\includegraphics[width=0.08\textwidth]{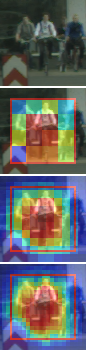}
\includegraphics[width=0.08\textwidth]{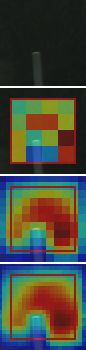}
\includegraphics[width=0.08\textwidth]{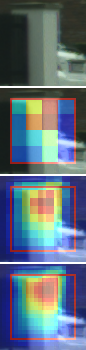}
\includegraphics[width=0.08\textwidth]{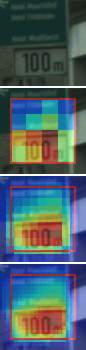}
\includegraphics[width=0.08\textwidth]{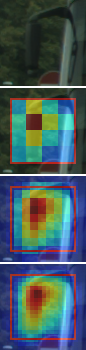}

\vspace{-3mm}
\caption{\footnotesize \textbf{Visualization of Hierarchical Attention on Cityscapes.} We show the original image in the top row, and our hierarchical attention with each row adding attention from a lower-level stage. The 2nd row shows attention from Stage 4,  the 3rd row shows Stage 3\&4, and the last row shows Stage 2\&3\&4. The red box represents receptive field for the Stage 4 features. Our attention mask captures the object boundary from higher-level to lower-level, showing the hierarchy from whole object into fine-detailed part masks, even for thin objects, such as poles. The last 3 examples are obtained from highlighted boxes in the first 3 examples in Figure 4 respectively.}
\label{figure:hierarchalattention}
\vspace{-5mm}
\end{figure}

We visualize the hierarchical attention mask in Figure~\ref{figure:hierarchalattention}. 
From Stage 4 to Stage 2, Our \method{} gradually  refines the semantic boundary of the whole object and is able to capture all types of categories from roads and skies to cars, people, and poles, even if the object is very narrow and small or severely occluded  by other objects.
This is also one of the reasons why we achieve significantly higher improvement in terms of boundary F-score for far-away objects, and we believe our hierarchical visualizations make it easier to understand the behaviour of high-level features in the hierarchical vision transformer. 

\vspace{-4mm}
\section{Conclusions}
\vspace{-3mm}
In this paper, we present \method{}, a lightweight attention-based method that extends hierarchical vision transformer architectures by adding local connections between features of different levels. We add \method{} to state-of-the-art methods in semantic segmentation and show greatly improved mIOU and F-score.  \method{} also creates highly informative hierarchical attention visualizations that aid in understanding the behaviour of higher-level features. In the future, we wish to explore more varied ways to propagate information across different levels.

\section{Acknowledgements}
This work was partially supported by NSERC. SF acknowledges the Canada CIFAR AI Chair award at the Vector Institute. GL, XZ, JG acknowledge support from the Vector Institute.


\clearpage
%
%
\bibliographystyle{splncs04}
\bibliography{egbib}

\begin{thebibliography}{10}
\providecommand{\url}[1]{\texttt{#1}}
\providecommand{\urlprefix}{URL }
\providecommand{\doi}[1]{https://doi.org/#1}

\bibitem{ba2016layernorm}
Ba, J.L., Kiros, J.R., Hinton, G.E.: Layer normalization. arXiv preprint
  arXiv:1607.06450  (2016)

\bibitem{bai2021visual}
Bai, S., Torr, P., et~al.: Visual parser: Representing part-whole hierarchies
  with transformers. arXiv preprint arXiv:2107.05790  (2021)

\bibitem{bau2017network}
Bau, D., Zhou, B., Khosla, A., Oliva, A., Torralba, A.: Network dissection:
  Quantifying interpretability of deep visual representations. In: Proceedings
  of the IEEE conference on computer vision and pattern recognition. pp.
  6541--6549 (2017)

\bibitem{hvt:chen2021regionvit}
Chen, C.F., Panda, R., Fan, Q.: Regionvit: Regional-to-local attention for
  vision transformers (2021)

\bibitem{chen2018deeplabv3+}
Chen, L.C., Zhu, Y., Papandreou, G., Schroff, F., Adam, H.: Encoder-decoder
  with atrous separable convolution for semantic image segmentation. In:
  Proceedings of the European conference on computer vision (ECCV). pp.
  801--818 (2018)

\bibitem{hvt:chen2021dpt}
Chen, Z., Zhu, Y., Zhao, C., Hu, G., Zeng, W., Wang, J., Tang, M.: Dpt:
  Deformable patch-based transformer for visual recognition. Proceedings of the
  29th ACM International Conference on Multimedia  (Oct 2021).
  \doi{10.1145/3474085.3475467},
  \url{http://dx.doi.org/10.1145/3474085.3475467}

\bibitem{cheng2021maskformer}
Cheng, B., Schwing, A.G., Kirillov, A.: Per-pixel classification is not all you
  need for semantic segmentation (2021)

\bibitem{hvt:chu2021twins}
Chu, X., Tian, Z., Wang, Y., Zhang, B., Ren, H., Wei, X., Xia, H., Shen, C.:
  Twins: Revisiting the design of spatial attention in vision transformers
  (2021)

\bibitem{mmseg2020}
Contributors, M.: {MMSegmentation}: Openmmlab semantic segmentation toolbox and
  benchmark. \url{https://github.com/open-mmlab/mmsegmentation} (2020)

\bibitem{cordts2016cityscapes}
Cordts, M., Omran, M., Ramos, S., Rehfeld, T., Enzweiler, M., Benenson, R.,
  Franke, U., Roth, S., Schiele, B.: The cityscapes dataset for semantic urban
  scene understanding. In: Proceedings of the IEEE conference on computer
  vision and pattern recognition. pp. 3213--3223 (2016)

\bibitem{hvt:dong2021cswin}
Dong, X., Bao, J., Chen, D., Zhang, W., Yu, N., Yuan, L., Chen, D., Guo, B.:
  Cswin transformer: A general vision transformer backbone with cross-shaped
  windows (2021)

\bibitem{dosovitskiy2020image}
Dosovitskiy, A., Beyer, L., Kolesnikov, A., Weissenborn, D., Zhai, X.,
  Unterthiner, T., Dehghani, M., Minderer, M., Heigold, G., Gelly, S., et~al.:
  An image is worth 16x16 words: Transformers for image recognition at scale.
  arXiv preprint arXiv:2010.11929  (2020)

\bibitem{dosovitskiy2020vit}
Dosovitskiy, A., Beyer, L., Kolesnikov, A., Weissenborn, D., Zhai, X.,
  Unterthiner, T., Dehghani, M., Minderer, M., Heigold, G., Gelly, S., et~al.:
  An image is worth 16x16 words: Transformers for image recognition at scale.
  arXiv preprint arXiv:2010.11929  (2020)

\bibitem{hvt:fan2021multiscale}
Fan, H., Xiong, B., Mangalam, K., Li, Y., Yan, Z., Malik, J., Feichtenhofer,
  C.: Multiscale vision transformers. In: Proceedings of the IEEE/CVF
  International Conference on Computer Vision. pp. 6824--6835 (2021)

\bibitem{fu2019danet}
Fu, J., Liu, J., Tian, H., Li, Y., Bao, Y., Fang, Z., Lu, H.: Dual attention
  network for scene segmentation. In: Proceedings of the IEEE/CVF conference on
  computer vision and pattern recognition. pp. 3146--3154 (2019)

\bibitem{gu2022hrvit}
Gu, J., Kwon, H., Wang, D., Ye, W., Li, M., Chen, Y.H., Lai, L., Chandra, V.,
  Pan, D.Z.: Multi-scale high-resolution vision transformer for semantic
  segmentation. In: Proceedings of the IEEE/CVF Conference on Computer Vision
  and Pattern Recognition. pp. 12094--12103 (2022)

\bibitem{he2016resnet}
He, K., Zhang, X., Ren, S., Sun, J.: Deep residual learning for image
  recognition. In: Proceedings of the IEEE conference on computer vision and
  pattern recognition. pp. 770--778 (2016)

\bibitem{hinton2021represent}
Hinton, G.: How to represent part-whole hierarchies in a neural network. arXiv
  preprint arXiv:2102.12627  (2021)

\bibitem{hinton2018matrix}
Hinton, G.E., Sabour, S., Frosst, N.: Matrix capsules with em routing. In:
  International conference on learning representations (2018)

\bibitem{huang2017densenet}
Huang, G., Liu, Z., Van Der~Maaten, L., Weinberger, K.Q.: Densely connected
  convolutional networks. In: Proceedings of the IEEE conference on computer
  vision and pattern recognition. pp. 4700--4708 (2017)

\bibitem{huang2019ccnet}
Huang, Z., Wang, X., Huang, L., Huang, C., Wei, Y., Liu, W.: Ccnet: Criss-cross
  attention for semantic segmentation. In: Proceedings of the IEEE/CVF
  International Conference on Computer Vision. pp. 603--612 (2019)

\bibitem{lecun1995convolutional}
LeCun, Y., Bengio, Y., et~al.: Convolutional networks for images, speech, and
  time series. The handbook of brain theory and neural networks
  \textbf{3361}(10), ~1995 (1995)

\bibitem{CelebAMask-HQ}
Lee, C.H., Liu, Z., Wu, L., Luo, P.: Maskgan: Towards diverse and interactive
  facial image manipulation. In: IEEE Conference on Computer Vision and Pattern
  Recognition (CVPR) (2020)

\bibitem{hvt:li2021localtoglobal}
Li, J., Yan, Y., Liao, S., Yang, X., Shao, L.: Local-to-global self-attention
  in vision transformers (2021)

\bibitem{ffn:li2021localvit}
Li, Y., Zhang, K., Cao, J., Timofte, R., Gool, L.V.: Localvit: Bringing
  locality to vision transformers (2021)

\bibitem{lin2019zigzagnet}
Lin, D., Shen, D., Shen, S., Ji, Y., Lischinski, D., Cohen-Or, D., Huang, H.:
  Zigzagnet: Fusing top-down and bottom-up context for object segmentation. In:
  Proceedings of the IEEE/CVF Conference on Computer Vision and Pattern
  Recognition. pp. 7490--7499 (2019)

\bibitem{lin2017fpn}
Lin, T.Y., Doll{\'a}r, P., Girshick, R., He, K., Hariharan, B., Belongie, S.:
  Feature pyramid networks for object detection. In: Proceedings of the IEEE
  conference on computer vision and pattern recognition. pp. 2117--2125 (2017)

\bibitem{liu2018panet}
Liu, S., Qi, L., Qin, H., Shi, J., Jia, J.: Path aggregation network for
  instance segmentation. In: Proceedings of the IEEE conference on computer
  vision and pattern recognition. pp. 8759--8768 (2018)

\bibitem{hvt:liu2021Swin}
Liu, Z., Lin, Y., Cao, Y., Hu, H., Wei, Y., Zhang, Z., Lin, S., Guo, B.: Swin
  transformer: Hierarchical vision transformer using shifted windows. In:
  Proceedings of the IEEE/CVF International Conference on Computer Vision. pp.
  10012--10022 (2021)

\bibitem{long2015fcn}
Long, J., Shelhamer, E., Darrell, T.: Fully convolutional networks for semantic
  segmentation. In: Proceedings of the IEEE conference on computer vision and
  pattern recognition. pp. 3431--3440 (2015)

\bibitem{park2019SPADE}
Park, T., Liu, M.Y., Wang, T.C., Zhu, J.Y.: Semantic image synthesis with
  spatially-adaptive normalization. In: Proceedings of the IEEE Conference on
  Computer Vision and Pattern Recognition (2019)

\bibitem{davisPerazzi2016}
Perazzi, F., Pont-Tuset, J., McWilliams, B., {Van Gool}, L., Gross, M.,
  Sorkine-Hornung, A.: A benchmark dataset and evaluation methodology for video
  object segmentation. In: Computer Vision and Pattern Recognition (2016)

\bibitem{ronneberger2015unet}
Ronneberger, O., Fischer, P., Brox, T.: U-net: Convolutional networks for
  biomedical image segmentation. In: International Conference on Medical image
  computing and computer-assisted intervention. pp. 234--241. Springer (2015)

\bibitem{sabour2017dynamic}
Sabour, S., Frosst, N., Hinton, G.E.: Dynamic routing between capsules.
  Advances in neural information processing systems  \textbf{30} (2017)

\bibitem{sandler2018mobilenetv2}
Sandler, M., Howard, A., Zhu, M., Zhmoginov, A., Chen, L.C.: Mobilenetv2:
  Inverted residuals and linear bottlenecks. In: Proceedings of the IEEE
  conference on computer vision and pattern recognition. pp. 4510--4520 (2018)

\bibitem{siam2017deepauto}
Siam, M., Elkerdawy, S., Jagersand, M., Yogamani, S.: Deep semantic
  segmentation for automated driving: Taxonomy, roadmap and challenges. In:
  2017 IEEE 20th international conference on intelligent transportation systems
  (ITSC). pp.~1--8. IEEE (2017)

\bibitem{sun2019hrnet}
Sun, K., Zhao, Y., Jiang, B., Cheng, T., Xiao, B., Liu, D., Mu, Y., Wang, X.,
  Liu, W., Wang, J.: High-resolution representations for labeling pixels and
  regions. arXiv preprint arXiv:1904.04514  (2019)

\bibitem{takahashi2021d3net}
Takahashi, N., Mitsufuji, Y.: Densely connected multi-dilated convolutional
  networks for dense prediction tasks. In: Proceedings of the IEEE/CVF
  Conference on Computer Vision and Pattern Recognition. pp. 993--1002 (2021)

\bibitem{takikawa2019gatedscnn}
Takikawa, T., Acuna, D., Jampani, V., Fidler, S.: Gated-scnn: Gated shape cnns
  for semantic segmentation. In: Proceedings of the IEEE/CVF international
  conference on computer vision. pp. 5229--5238 (2019)

\bibitem{takikawa2019gated}
Takikawa, T., Acuna, D., Jampani, V., Fidler, S.: Gated-scnn: Gated shape cnns
  for semantic segmentation. In: Proceedings of the IEEE/CVF international
  conference on computer vision. pp. 5229--5238 (2019)

\bibitem{teichmann123autonomous}
Teichmann123, M., Weber, M., Z{\"o}llner, M., Cipolla, R., Urtasun, R.:
  Multinet: Real-time joint semantic reasoning for autonomous driving

\bibitem{vaswani2017attention}
Vaswani, A., Shazeer, N., Parmar, N., Uszkoreit, J., Jones, L., Gomez, A.N.,
  Kaiser, {\L}., Polosukhin, I.: Attention is all you need. Advances in neural
  information processing systems  \textbf{30} (2017)

\bibitem{hvt:wang2021pyramid}
Wang, W., Xie, E., Li, X., Fan, D.P., Song, K., Liang, D., Lu, T., Luo, P.,
  Shao, L.: Pyramid vision transformer: A versatile backbone for dense
  prediction without convolutions. In: Proceedings of the IEEE/CVF
  International Conference on Computer Vision. pp. 568--578 (2021)

\bibitem{hvt:wang2021pvtv2}
Wang, W., Xie, E., Li, X., Fan, D.P., Song, K., Liang, D., Lu, T., Luo, P.,
  Shao, L.: Pvtv2: Improved baselines with pyramid vision transformer.
  Computational Visual Media  \textbf{8}(3),  1--10 (2022)

\bibitem{hvt:scale:wang2021crossformer}
Wang, W., Yao, L., Chen, L., Lin, B., Cai, D., He, X., Liu, W.: Crossformer: A
  versatile vision transformer hinging on cross-scale attention (2021)

\bibitem{wang2018non}
Wang, X., Girshick, R., Gupta, A., He, K.: Non-local neural networks. In:
  Proceedings of the IEEE conference on computer vision and pattern
  recognition. pp. 7794--7803 (2018)

\bibitem{hvt::wu2021cvt}
Wu, H., Xiao, B., Codella, N., Liu, M., Dai, X., Yuan, L., Zhang, L.: Cvt:
  Introducing convolutions to vision transformers (2021)

\bibitem{hvt:wu2021pale}
Wu, S., Wu, T., Tan, H., Guo, G.: Pale transformer: A general vision
  transformer backbone with pale-shaped attention (2021)

\bibitem{hvt:wu2021p2t}
Wu, Y.H., Liu, Y., Zhan, X., Cheng, M.M.: P2t: Pyramid pooling transformer for
  scene understanding. arXiv preprint arXiv:2106.12011  (2021)

\bibitem{xiao2018upernet}
Xiao, T., Liu, Y., Zhou, B., Jiang, Y., Sun, J.: Unified perceptual parsing for
  scene understanding. In: Proceedings of the European Conference on Computer
  Vision (ECCV). pp. 418--434 (2018)

\bibitem{hvt:xie2021segformer}
Xie, E., Wang, W., Yu, Z., Anandkumar, A., Alvarez, J.M., Luo, P.: Segformer:
  Simple and efficient design for semantic segmentation with transformers.
  Advances in Neural Information Processing Systems  \textbf{34} (2021)

\bibitem{hvt:yang2021focal}
Yang, J., Li, C., Zhang, P., Dai, X., Xiao, B., Yuan, L., Gao, J.: Focal
  self-attention for local-global interactions in vision transformers (2021)

\bibitem{yu2015multi}
Yu, F., Koltun, V.: Multi-scale context aggregation by dilated convolutions.
  arXiv preprint arXiv:1511.07122  (2015)

\bibitem{yu2018dla}
Yu, F., Wang, D., Shelhamer, E., Darrell, T.: Deep layer aggregation. In:
  Proceedings of the IEEE conference on computer vision and pattern
  recognition. pp. 2403--2412 (2018)

\bibitem{yuan2020ocrnet}
Yuan, Y., Chen, X., Wang, J.: Object-contextual representations for semantic
  segmentation. In: European conference on computer vision. pp. 173--190.
  Springer (2020)

\bibitem{zagoruyko2016wideresnet}
Zagoruyko, S., Komodakis, N.: Wide residual networks. arXiv preprint
  arXiv:1605.07146  (2016)

\bibitem{zeiler2014visualizing}
Zeiler, M.D., Fergus, R.: Visualizing and understanding convolutional networks.
  In: European conference on computer vision. pp. 818--833. Springer (2014)

\bibitem{zhang2020fpt}
Zhang, D., Zhang, H., Tang, J., Wang, M., Hua, X., Sun, Q.: Feature pyramid
  transformer. In: European Conference on Computer Vision. pp. 323--339.
  Springer (2020)

\bibitem{zhao2021graphfpn}
Zhao, G., Ge, W., Yu, Y.: Graphfpn: Graph feature pyramid network for object
  detection. In: Proceedings of the IEEE/CVF International Conference on
  Computer Vision. pp. 2763--2772 (2021)

\bibitem{zhao2017pspnet}
Zhao, H., Shi, J., Qi, X., Wang, X., Jia, J.: Pyramid scene parsing network.
  In: Proceedings of the IEEE conference on computer vision and pattern
  recognition. pp. 2881--2890 (2017)

\bibitem{ade20k}
Zhou, B., Zhao, H., Puig, X., Fidler, S., Barriuso, A., Torralba, A.: Scene
  parsing through ade20k dataset. In: 2017 IEEE Conference on Computer Vision
  and Pattern Recognition (CVPR). pp. 5122--5130 (2017).
  \doi{10.1109/CVPR.2017.544}

\end{thebibliography}
\clearpage

\appendix
\subtitle{Supplementary Material}
In the supplementary material, we provide more details about our model \method{} in Section~\ref{sec:model_details}, and experimental settings in Section~\ref{sec:exp_details}. Further ablation studies are provided in Section~\ref{sec:more_ablations}, with more qualitative examples on both Cityscapes and ADE20K datasets in Section~\ref{sec:more_seg_results}.

\section{Model Details}
\label{sec:model_details}
We provide further details about our inter-level attention layer, the efficient implementation of top-down updates,  hierarchical visualization, as well as the complexity analysis of our models. 

\subsection{Inter-Level Attention Layer}
Our attention layer follows the design of standard self-attention blocks~\cite{hvt:liu2021Swin,hvt:yang2021focal,hvt:wang2021pvtv2} and each layer is split into two steps of Attention and the Feed-Forward Network (FFN). We show a generic case of our attention layer below with example vectors $X_1 \in \mathbb{R}^{n_1 \times d_{1}}$ and $X_2 \in \mathbb{R}^{n_2 \times d_{2}}$, and we remove the subscript and superscript from the main paper for simplicity. 
\begin{eqnarray}
X^{\prime}_{1}&=& X^{}_{1} +  \text{attn}(X^{}_{1}, X^{}_{2}),\\ 
X^{\prime\prime}_{1} &=&\alpha X^{\prime}_{1} + \beta \text{FFN}(X^{\prime}_{2}),
\end{eqnarray} 
where hyper-parameters $\alpha$ and $\beta$ control how much information is updated by the feed-forward network.

\paragraph{Attention} We use the standard attention framework in HVT architectures. We first apply layer norm \cite{ba2016layernorm} into $X_1$ and $X_2$. We then convert our inputs into the query, key, and value by a separate fully-connected layers $q, k , v$ respectively. The feature we wish to update, $X_1$, will be converted into the query $Q = q(X_1) \in \mathbb{R}^{n_1\times d}$ and our attending features $X_2$ will be converted to the key and value $K, V = k(X_{2}), v(X_2) \in \mathbb{R}^{n_2\times d}$. For our attention update, we need to unify the dimensions of the two inputs. In our case, we always use the lower dimension $d = \text{min}(d_1, d_2)$ to reduce complexity. Our dot product attention is then calculated using:
\begin{eqnarray}
M = \text{Softmax}(\frac{QK^{\top}}{\sqrt{d}} + B), \\
\text{attn}(X_1, X_2) = f(MV),
\end{eqnarray}
where $M$ denotes the intermediate attention weight matrix and $f$ is the output fully connected layer which projects our attention output to the dimension of the vector we are updating, with our case being $\mathbb{R}^{d} \xrightarrow{} \mathbb{R}^{d_1}$. We do not use multi-headed attention to ensure that there is only one clear set of attention weights between features of different levels. With multi-headed attention, it is possible that different heads of the attention will attend to different parts of the local patch area, resulting in more ``mixing'' of information which is not ideal.

\paragraph{Feed-Forward Network} We use Segformer's~\cite{hvt:xie2021segformer} FFN design, which is shown below:
\begin{equation}
X_{out} =\text{FFN}(X_{in}) = \text{MLP}(\text{GELU}(\text{Conv}_{3\times3}(\text{MLP}(LN(X_{in})))),
\end{equation}
where LN stands for Layer Norm~\cite{ba2016layernorm}, and Conv stands for a depth-wise convolution. 

\subsection{Implementation of Top-Down Inter-Level Attention}

In our Top-Down Inter-Level Attention, we attend to lower-level features $X^{}_{\lo{}}$ using higher-level features $X^{}_{\hi{}, \hat{P}_{}}$, where $\hat{P}$ denotes locations of higher-level features whose local patches cover the lower-level features.
Programming-wise, we can obtain the local patches $P$ easily through Pytorch's F.unfold operation, which separates an input into patches based on kernel size and stride. However it is not trivial to invert our local patches $P$ in an efficient manner to obtain $\hat{P}$ and then use $\hat{P}$ in attention. 

This is because the number of higher-level features in $\hat{P}$ can range from 1 to 4 depending on the lower-level feature's position, with edges having less, which results in large difficulty in constructing features in tensor format. We use a clever trick to circumvent these issues. 
Instead of directly computing our top-down attention with $QK^T$, where $Q \in \mathbb{R}^{1\times d}$ and $K\in \mathbb{R}^{4\times d}$, we  compute it
within a local patch $P$, where $Q \in \mathbb{R}^{16\times d}$ and $K\in \mathbb{R}^{1\times d}$  for all of the high-level features, and utilize the operations from PyTorch to map it back and make sure they are mathematically equivalent. 

Intuitively, we know that each higher-level feature will participate in the Top-Down Inter-Level Attention for all lower-level features in it's local patch. As such, we can first calculate the individual weights across the local patch and then run the softmax across $\hat{P}_{}$ later, circumventing our edge issue. This is mathematically equivalent to $\text{attn}(X^{}_{\lo{}}, X^{}_{\hi{}, \hat{P}_{}})$. We implement our softmax across $\hat{P}_{}$ by using Pytorch's F.fold operation on the intermediate output before our softmax. This collapses every patch's attention weights into the original image size, with overlapping patch values summing together across $\hat{P}_{}$. We then separate this summed weights into patches again using F.unfold and then use this summed value for the softmax operation. We show our code\footnote{Our implementation uses Pytorch 1.7.1} below:

\begin{lstlisting}[language=Python, caption=Code of Top-Down and Bottom-Up Inter-Level Attention Implementation]
import torch.nn.functional as F

def top_down_interlevel_attention(Xt, Xb):
    # Xt: higher-level features, shape is (B, dt, Ht, Wt)
    # Xb: bottom-level features, shape is (B, db, Hb, Wb)
    
    # Get local patches for each higher-level feature
    Xbp = F.unfold(Xb, kernel_size=4, stride=2, padding=1)  
    Xbp = Xbp.permute(0, 2, 1).reshape(B, Ht * Wt, 16, db)
    
    # q, k, v are fully connected layers
    Xt = Xt.permute(0, 2, 3, 1).reshape(B, Ht * Wt, 1, dt)
    Q, K, V = q(Xbp), k(Xt), v(Xt)  
    
    # @ is the matmul operation, B is relative coordinates 
    # f is output fully connected layer
    M = softmax_P(Q @ K.transpose(3, 4) / db ** 0.5 + B)
    top_down_f = f(M @ V).permute(0, 3, 2, 1) 
    top_down_f = top_down_f.reshape(B, 16 * db, Ht * Wt)
    
    # Sum attention values across local patches
    top_down_f = F.fold(top_down_f, output_size=(Hb, Wb), 
        kernel_size=4, stride=2, padding=1)    
    return top_down_f

def softmax_P(x):
    x = torch.exp(x) 
    
    # Sum weights across overlapping patches
    attn = attn.permute(0,2,3,1).reshape(B, 16, Ht * Wt)
    attn_sum = F.fold(attn, output_size=(Hb, Wb), 
        kernel_size=4, stride=2, padding=1) # (B, 1, Hb, Wb)
    
    # Unfold the sum into patches
    attn_sum_p = F.unfold(attn_sum, kernel_size=4, stride=2, 
        padding=1).permute(0, 2, 1)  # (B, Ht * Wt, 16)
    
    attn_sum_p[attn_sum_p == 0] = 1.0
    return x / attn_sum_p.reshape(B, Ht * Wt, 16, 1)

def bottom_up_interlevel_attention(Xt, Xb):
    # Xt: higher-level features, shape is (B, dt, Ht, Wt)
    # Xb: bottom-level features, shape is (B, db, Hb, Wb)
    
    # Get local patches for each higher-level feature
    Xbp = F.unfold(Xb, kernel_size=4, stride=2, padding=1)  
    Xbp = Xbp.permute(0, 2, 1).reshape(B, Ht * Wt, 16, db)
    
    # q, k, v are fully connected layers
    Xt = Xt.permute(0, 2, 3, 1).reshape(B, Ht * Wt, 1, dt)
    Q, K, V = q(Xt), k(Xbp), v(Xbp)  
    
    # @ is the matmul operation, B is relative coordinates 
    # f is output fully connected layer
    M = softmax(Q @ K.transpose(3, 4) / db ** 0.5 + B)
    bottom_up_f = f(M @ V).permute(0, 3, 2, 1) 
    return bottom_up_f
\end{lstlisting}

\subsection{Hierarchical Visualization}

During our Top-Down Attention we obtain the attention weights $M^{} \in \mathbb{R}^{H_{\hi{}}W_{\hi{}} \times 16 \times 1}$ across multiple stages. For a specific higher-level feature at a stage $\ell$ and position $\{h_{\hi{}},w_{\hi{}}\}$, the attention $M^{}_{\ell, \{h_{\hi{}},w_{\hi{}}\}} \in \mathbb{R}^{16\times 1}$ represents how much each lower-level feature within it's local patch aligns with it compared to other overlapping higher-level features. We can normalize this value at Stage 4 to obtain our hierarchical visualization at its most coarse representation of one single layer between Stage 3 and 4. 
For equations from this point, we use $\{h_{\hi{}},w_{\hi{}}\}$ for indexing.  We denote the assignment between stage $\ell_1$ and stage $\ell_0$ as $M^{}_{\ell_1 \to \ell_0}$ and note that $M^{}_{\ell} = M^{}_{\ell \to \ell -1}$.

Each Stage 3 feature also has the corresponding assignments to its lower-level features in Stage 2, $M^{}_{3 \to 2}$. We can further build upon these attention assignments to compute the assignment from stage 4 to stage 2. Specifically, we iterate all the intermediate stage 3 attention matrices, and for each Stage 3 attention matrix, $M^{}_{3\to2, \{h_3,w_3\},\{h_2,w_2\}}$, we can weigh it using the specific attention value from Stage 4, $M^{}_{4\to3, \{h_4,w_4\}, \{h_3,w_3\}}$. We multiply the values together and sum the weights across overlapping patches before normalizing. Our new Stage 2 to Stage 4 attention weights is then:
\begin{eqnarray}
M^{}_{4\to 2, \{h_4,w_4\},\{h_2,w_2\}} &=&  \sum_{\{h_3,w_3\} \in P_4}M^{}_{4\to 3, \{h_4,w_4\}, \{h_3,w_3\}}* M^{}_{3\to 2, \{h_3,w_3\}, \{h_2,w_2\}}, \\
\bar{M}^{}_{4\to 2, \{h_4,w_4\}\{h_2,w_2\}} &=& \frac{M^{}_{4\to 2, \{h_4,w_4\},\{h_2,w_2\}}}{\sum_{\{h_2,w_2\}} M^{}_{4\to 2,\{h_4,w_4\},\{h_2,w_2\}}},
\end{eqnarray}
with our full Stage 1 to Stage 4 attention weights being:
\begin{eqnarray}
M^{}_{4\to 1, \{h_4,w_4\},\{h_1,w_1\}} &=&  \sum_{\{h_2,w_2\} \in P_3}M^{}_{4\to 2, \{h_4,w_4\}, \{h_2,w_2\}}* M^{}_{2\to 1, \{h_2,w_2\},\{h_1,w_1\}}, \\
\bar{M}^{}_{4\to 1, \{h_4,w_4\},\{h_1,w_1\}} &=& \frac{M^{}_{4\to 1, \{h_4,w_4\},\{h_1,w_1\}}}{\sum_{\{h_1,w_1\}} M^{}_{4\to 1, \{h_4,w_4\},\{h_1,w_1\}}},
\end{eqnarray}
We use these normalized these attention matrices for our visualizations.

\subsection{Complexity Analysis}
We show that our Inter-Level Attention is computationally efficient to compute. We attend between higher-level features $X_\hi{} \in \mathbb{R}^{H_\hi{}\times W_\hi{}\times d_\hi{}}$ and lower-level features $X_{\lo{}} \in \mathbb{R}^{H_{\lo{}}\times W_{\lo{}}\times d_{\lo{}}}$ within a local window $P$, where our kernel of (4 by 4) results in 16 elements in P. We go over the two components in our dot-product attention: the fully connected layers and the dot-product operation. 

\paragraph{Fully Connected Layers:} There are 4 linear projection operators in dot-product attention, where 3 come from converting our inputs into the query, key and value, and 1 from the final layer which projects our attention output. The complexity of a fully connected layer which projects from $\mathbb{R}^{HW\times d_1} \xrightarrow{} \mathbb{R}^{HW\times d_2}$ is $HWd_1d_2$. We work with different dimensions between higher-level and lower-level features and need to unify dimensions for the dot-product attention step. We always project dimensions downwards to those of the lower-level features to reduce computation, before projecting back upwards again if required in the final projection layer. We lay out the complexities of our updates below in Table~\ref{table:fullyconnectedcomplexity}, with the overall complexity for both Top-Down and Bottom-Update being:

\begin{equation}
\Omega(\text{Fully Connected Layers}) = 2H_\hi{}W_\hi{}d_\hi{}d_{\lo{}} + 2W_{\lo{}}W_{\lo{}}d_{\lo{}}^2
\end{equation} 

\begin{table}[]
\caption{\textbf{Complexity of our Fully Connected Layers}}
\begin{adjustbox}{width=\textwidth}
\begin{tabular}{|c|cccc|cccc|}
\hline
 &
  \multicolumn{4}{c|}{\textbf{Bottom-Up Attention}} &
  \multicolumn{4}{c|}{\textbf{Top-Down Attention}} \\ \hline
 &
  \multicolumn{1}{c|}{Query} &
  \multicolumn{1}{c|}{Key} &
  \multicolumn{1}{c|}{Value} &
  Final Projection &
  \multicolumn{1}{c|}{Query} &
  \multicolumn{1}{c|}{Key} &
  \multicolumn{1}{c|}{Value} &
  Final Projection \\ \hline
\textbf{Operation} &
  \multicolumn{1}{c|}{$q(X^{i}_{\hi{}})$} &
  \multicolumn{1}{c|}{$k(X^{i}_{\lo{}})$} &
  \multicolumn{1}{c|}{$v(X^{i}_{\lo{}})$} &
  \multicolumn{1}{c|}{$\text{}f(X^{i}_{\hi{}},X^{i}_{\lo{}})$} &
  \multicolumn{1}{c|}{$q(X^{i}_{\lo{}})$} &
  \multicolumn{1}{c|}{$k(X^{i}_{\hi{}})$} &
  \multicolumn{1}{c|}{$v(X^{i}_{\hi{}})$} &
  \multicolumn{1}{c|}{$\text{}f(X^{i}_{\lo{}},X^{i}_{\hi{}})$} 
   \\ \hline
\textbf{Input} &
  \multicolumn{1}{c|}{\begin{tabular}[c]{@{}c@{}}$\mathbb{R}^{H_{\hi{}}W_{\hi{}} \times d_{\hi{}}}$\\  \end{tabular}} &
  \multicolumn{1}{c|}{\begin{tabular}[c]{@{}c@{}}$\mathbb{R}^{H_{\lo{}}W_{\lo{}} \times d_{\lo{}}}$\\  \end{tabular}} &
  \multicolumn{1}{c|}{\begin{tabular}[c]{@{}c@{}}$\mathbb{R}^{H_{\lo{}}W_{\lo{}} \times d_{\lo{}}}$\\  \end{tabular}} &
  \multicolumn{1}{c|}{\begin{tabular}[c]{@{}c@{}}$\mathbb{R}^{H_{\hi{}}W_{\hi{}} \times d_{\lo{}}}$\\  \end{tabular}} &
  \multicolumn{1}{c|}{\begin{tabular}[c]{@{}c@{}}$\mathbb{R}^{H_{\lo{}}W_{\lo{}} \times d_{\lo{}}}$\\  \end{tabular}} &
  \multicolumn{1}{c|}{\begin{tabular}[c]{@{}c@{}}$\mathbb{R}^{H_{\hi{}}W_{\hi{}} \times d_{\hi{}}}$\\  \end{tabular}} &
  \multicolumn{1}{c|}{\begin{tabular}[c]{@{}c@{}}$\mathbb{R}^{H_{\hi{}}W_{\hi{}} \times d_{\hi{}}}$\\  \end{tabular}} &
  \multicolumn{1}{c|}{\begin{tabular}[c]{@{}c@{}}$\mathbb{R}^{H_{\lo{}}W_{\lo{}} \times d_{\lo{}}}$\\  \end{tabular}} 
   \\ \hline
\textbf{Output} &
  \multicolumn{1}{c|}{$\mathbb{R}^{H_{\hi{}} \times W_{\hi{}} \times d_{\lo{}}}$} &
  \multicolumn{1}{c|}{\begin{tabular}[c]{@{}c@{}}$\mathbb{R}^{H_{\lo{}}W_{\lo{}} \times d_{\lo{}}}$\\  \end{tabular}} &
  \multicolumn{1}{c|}{\begin{tabular}[c]{@{}c@{}}$\mathbb{R}^{H_{\lo{}}W_{\lo{}} \times d_{\lo{}}}$\\  \end{tabular}} &
  \multicolumn{1}{c|}{\begin{tabular}[c]{@{}c@{}}$\mathbb{R}^{H_{\hi{}}W_{\hi{}} \times d_{\hi{}}}$\\  \end{tabular}} &
  \multicolumn{1}{c|}{\begin{tabular}[c]{@{}c@{}}$\mathbb{R}^{H_{\lo{}}W_{\lo{}} \times d_{\lo{}}}$\\  \end{tabular}} &
  \multicolumn{1}{c|}{\begin{tabular}[c]{@{}c@{}}$\mathbb{R}^{H_{\hi{}}W_{\hi{}} \times d_{\lo{}}}$\\  \end{tabular}} &
  \multicolumn{1}{c|}{\begin{tabular}[c]{@{}c@{}}$\mathbb{R}^{H_{\hi{}}W_{\hi{}} \times d_{\lo{}}}$\\  \end{tabular}} &
  \multicolumn{1}{c|}{\begin{tabular}[c]{@{}c@{}}$\mathbb{R}^{H_{\lo{}}W_{\lo{}} \times d_{\lo{}}}$\\  \end{tabular}} 
   \\ \hline
\textbf{Complexity} &
  \multicolumn{1}{c|}{$H_\hi{}W_\hi{}d_\hi{}d_{\lo{}}$} &
  \multicolumn{1}{c|}{$H_{\lo{}}W_{\lo{}}d_{\lo{}}^2$} &
  \multicolumn{1}{c|}{$H_{\lo{}}W_{\lo{}}d_{\lo{}}^2$} &
   $H_\hi{}W_\hi{}d_\hi{}d_{\lo{}}$ &
  \multicolumn{1}{c|}{$H_{\lo{}}W_{\lo{}}d_{\lo{}}^2$} &
  \multicolumn{1}{c|}{$H_\hi{}W_\hi{}d_\hi{}d_{\lo{}}$} &
  \multicolumn{1}{c|}{$H_\hi{}W_\hi{}d_\hi{}d_{\lo{}}$} &
  $H_{\lo{}}W_{\lo{}}d_{\lo{}}^2$ \\ \hline
\textbf{Total} &
  \multicolumn{4}{c|}{$2H_\hi{}W_\hi{}d_\hi{}d_{\lo{}} + 2H_{\lo{}}W_{\lo{}}d_{\lo{}}^2$} &
  \multicolumn{4}{c|}{$2H_\hi{}W_\hi{}d_\hi{}d_{\lo{}} + 2H_{\lo{}}W_{\lo{}}d_{\lo{}}^2$} \\ \hline
\end{tabular}
\end{adjustbox}
\label{table:fullyconnectedcomplexity}
\end{table}

\paragraph{Dot-Product Attention:} The complexity of the dot product attention between vectors $Q \in \mathbb{R}^{HW \times N_1\times d}$ and $K, V \in \mathbb{R}^{HW \times N_2\times d}$ in the attention operation 
\begin{equation}
\label{eqn:selfattention}
\text{attn}(X_1, X_2) = \text{Softmax}(\frac{QK^{\top}}{\sqrt{d_\ell}} + B)V, 
\end{equation}
is $2dHWN_1N_2$. In our Bottom-Up Attention, $X_1 = q(X_\hi{}) \in \mathbb{R}^{H_\hi{}W_\hi{}\times 1 \times d_{\lo{}}}$ and $X_2 = k(X_{\lo{}}), v(X_{\lo{}}) \in \mathbb{R}^{H_\hi{}W_\hi{}\times16\times d_\lo{}}$, where our dimension of 16 is due our attention being limited to our local patch $P$ of size (4, 4). As such the complexity is $32d_{\lo{}}H_\hi{}W_\hi{}$. In our Top-Down Attention, $X_1 = q(X_{\lo{}}) \in \mathbb{R}^{H_{\lo{}}W_{\lo{}}\times1\times d_{\lo{}}}$ and $X_2 = k(X_{\hi{}}), v(X_{\hi{}}) \in \mathbb{R}^{H_\lo{}W_\lo{}\times4\times d_\lo{}}$, where our dimension of 4 is due each lower-level feature being attended by 4 higher-level features. As such the complexity is $8d_{\lo{}}H_{\lo{}}W_{\lo{}}$. As our HVT architectures downsample by a factor of 2 in Stage 2 and onwards, $4H_{\lo{}}W_{\lo{}} = 16H_\hi{}W_\hi{}$. Our final dot-product complexity is then:   

\begin{equation}
\Omega(\text{Dot-Product Attention}) = 32d_{\lo{}}H_\hi{}W_\hi{}
\end{equation} 

\paragraph{Comparison to Self-Attention:} We compare the complexity of Self-Attention on features $X \in \mathbb{R}^{HW\times d}$ to our Inter-Level Attention:

\begin{equation}
\Omega(\text{Self-Attention}) = 4HWd^2 + 2dH^2W^2
\end{equation} 
\begin{equation}
\Omega(\text{Inter-Level Attention}) = 2H_\hi{}W_\hi{}d_\hi{}d_{\lo{}} + 2H_{\lo{}}W_{\lo{}}d_{\lo{}}^2 + 32d_{\lo{}}H_\hi{}W_\hi{}
\end{equation} 
Attention complexity mainly comes from the dot-product attention term as $HW >> d$. Compared to self-attention our local patch reduces a complexity of $dH^2W^2$ to $16dHW$. As such our attention operation is very computationally efficient, especially on large images.

\subsection{Example Figure of Model}

\begin{figure*}[t!]
\centering
\includegraphics[width=0.9\linewidth, trim=1cm 5cm 3cm 5cm,]{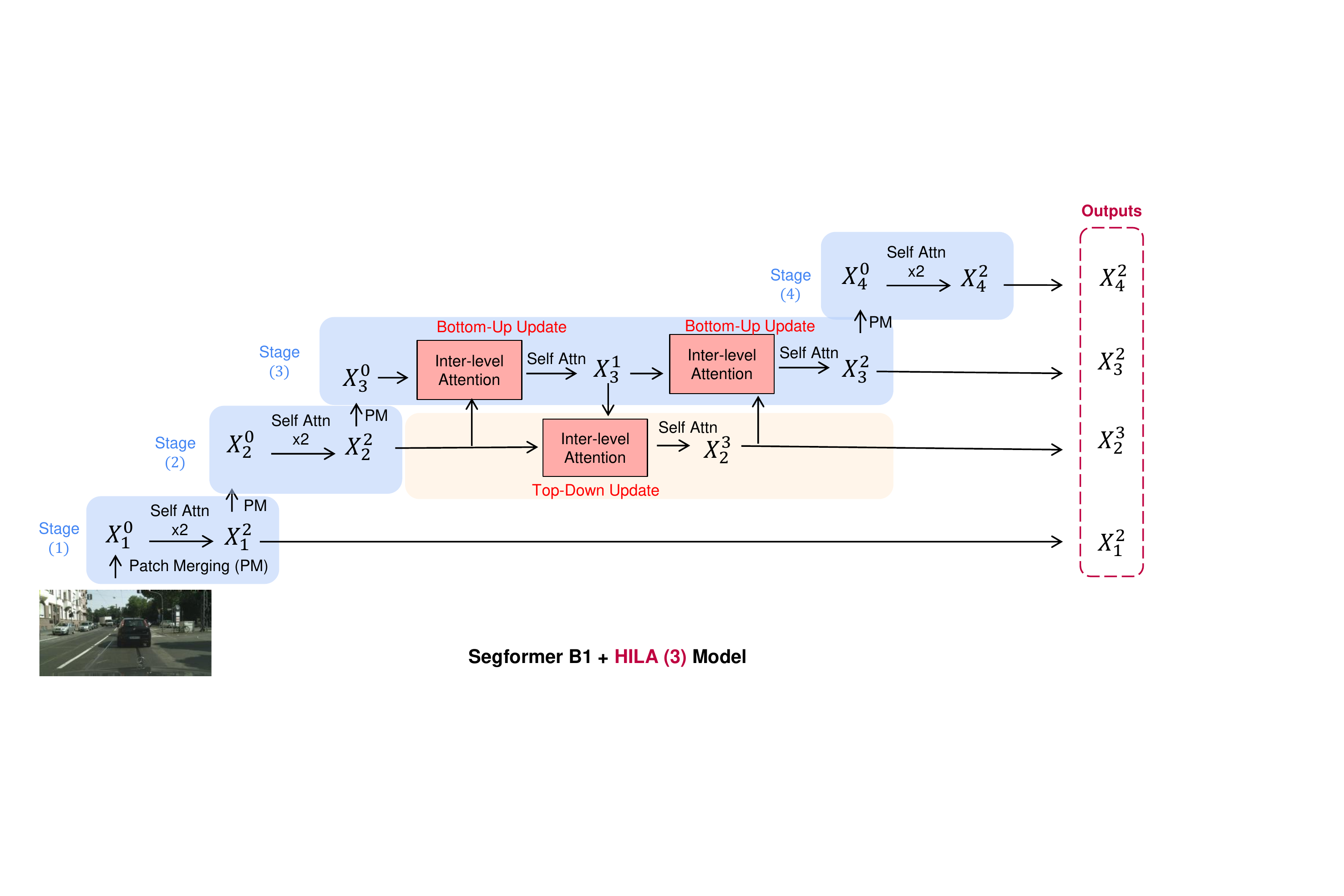}
\caption{\footnotesize \textbf{Full Diagram of SegFormer B1 + \methodone{}.} An example model, with \method applied to Stage 3. For our \methodthree{} models, Stage 2 and Stage 4 would have similar updates as Stage 3 in this case.}
\label{fig:full_model}
\end{figure*}

We provide a figure showing the entire SegFormer + \methodone{} model in Figure~\ref{fig:full_model}, as a more in-depth reference to how one of our models would look in its entirety.

\section{Experimental Details}
\label{sec:exp_details}
In this section, we provide more experimental details, including the hyper-parameters for our models, the source of compared baseline models, and our modified F-score for ADE20k. 
\subsection{Model Hyper-parameters}

We list the hyper-parameters for \method{} and the backbone models in Table~\ref{table:hyperparameters}. The Self-Attention Layer in \method{}'s Top-Down Update uses the same hyper-parameters as the original backbone's self-attention block. For the decoding head, we use the same parameters as the original model for all backbone models. 

\paragraph{General Hyper-Parameters}
The subscript $\ell$ denotes the stage of the hyper-parameter.
\begin{itemize}
    \item $K_\ell$: the patch merging convolution kernel size
    \item $S_\ell$: the patch merging convolution stride size
    \item $d_\ell$: the channel size of the stage
    \item $N_\ell$: the number of blocks in the stage
    \item $H_\ell$: the number of heads used in attention
    \item $E_\ell$: the feed-forward layer expansion ratio
\end{itemize}

\paragraph{HILA-Specific Hyper-Parameters}
\begin{itemize}
    \item $\alpha_\ell, \beta_\ell$: information propagation constants
    \item $s_\ell$: the stride at which \method{} is applied. For deeper models, we apply \method{} every $s_\ell$ blocks in one stage.
    \item $p_\ell$: the local patch size between higher and lower-level features
\end{itemize}

\paragraph{Segformer-Specific Hyper-Parameters}
\begin{itemize}
    \item $R_\ell$: the reduction ratio in spatial reduction attention
\end{itemize}

\paragraph{Swin Transformer-Specific Hyper-Parameters}
\begin{itemize}
    \item $W_\ell$: the local shifted window size
\end{itemize}

\begin{table}[]
\begin{center}
\caption{\textbf{Detailed Hyper-Parameters of \method{} models}}
\begin{adjustbox}{width=\textwidth}
\begin{tabular}{|c|c|c|cccc|c|}
\hline
\multirow{2}{*}{} &
  \multirow{2}{*}{Output Size} &
  \multirow{2}{*}{Component} &
  \multicolumn{4}{c|}{Segformer} &
  Swin \\ \cline{4-8} 
 &
   &
   &
  \multicolumn{1}{c|}{\begin{tabular}[c]{@{}c@{}}B0 +\\ \methodone{}\end{tabular}} &
  \multicolumn{1}{c|}{\begin{tabular}[c]{@{}c@{}}B1 +\\ \methodone{}\end{tabular}} &
  \multicolumn{1}{c|}{\begin{tabular}[c]{@{}c@{}}B1 +\\ \methodthree{}\end{tabular}} &
  \begin{tabular}[c]{@{}c@{}}B2 + \\ \methodthree{}\end{tabular} &
  \begin{tabular}[c]{@{}c@{}}Tiny +\\ \methodthree{}\end{tabular} \\ \hline
\multirow{2}{*}{Stage 1} &
  \multirow{2}{*}{$\frac{H}{4} \times \frac{W}{4}$} &
  Patch Merging &
   \multicolumn{1}{c|}{\begin{tabular}[c]{@{}c@{}}$K_1$=7 \\ $S_1$=4 \\ $d_1$=32\end{tabular}} &
  \multicolumn{1}{c|}{\begin{tabular}[c]{@{}c@{}}$K_1$=7 \\ $S_1$=4 \\ $d_1$=64\end{tabular}} &
  \multicolumn{1}{c|}{\begin{tabular}[c]{@{}c@{}}$K_1$=7 \\ $S_1$=4 \\ $d_1$=64\end{tabular}} &
  \multicolumn{1}{c|}{\begin{tabular}[c]{@{}c@{}}$K_1$=7 \\ $S_1$=4 \\ $d_1$=64\end{tabular}} &
  \multicolumn{1}{c|}{\begin{tabular}[c]{@{}c@{}}$K_1$=4 \\ $S_1$=4 \\ $d_1$=96\end{tabular}}
   \\ \cline{3-8} 
  &
   &
  \begin{tabular}[c]{@{}c@{}}Self-Attention \\ Layer\end{tabular} &
  \multicolumn{1}{c|}{\begin{tabular}[c]{@{}c@{}}$R_1$=8 \\ $H_1$=1 \\ $E_1$=4  \\ $N_1$=2 \end{tabular}} &
  \multicolumn{1}{c|}{\begin{tabular}[c]{@{}c@{}}$R_1$=8 \\ $H_1$=1 \\ $E_1$=4  \\ $N_1$=2 \end{tabular}} &
  \multicolumn{1}{c|}{\begin{tabular}[c]{@{}c@{}}$R_1$=8 \\ $H_1$=1 \\ $E_1$=4  \\ $N_1$=2 \end{tabular}} &
  \multicolumn{1}{c|}{\begin{tabular}[c]{@{}c@{}}$R_1$=8 \\ $H_1$=1 \\ $E_1$=4  \\ $N_1$=2 \end{tabular}} &
  \multicolumn{1}{c|}{\begin{tabular}[c]{@{}c@{}}$W_1$=7 \\ $H_1$=3 \\ $E_1$=4  \\ $N_1$=2 \end{tabular}}
   \\ \hline
\multirow{3}{*}{Stage 2} &
  \multirow{3}{*}{$\frac{H}{8} \times \frac{W}{8}$} &
  Patch Merging &
  \multicolumn{1}{c|}{\begin{tabular}[c]{@{}c@{}}$K_2$=3 \\ $S_2$=2 \\ $d_2$=64\end{tabular}} &
  \multicolumn{1}{c|}{\begin{tabular}[c]{@{}c@{}}$K_2$=3 \\ $S_2$=2 \\ $d_2$=128\end{tabular}} &
  \multicolumn{1}{c|}{\begin{tabular}[c]{@{}c@{}}$K_2$=3 \\ $S_2$=2 \\ $d_2$=128\end{tabular}} &
   \multicolumn{1}{c|}{\begin{tabular}[c]{@{}c@{}}$K_2$=3 \\ $S_2$=2 \\ $d_2$=128\end{tabular}} &
   \multicolumn{1}{c|}{\begin{tabular}[c]{@{}c@{}}$K_2$=2 \\ $S_2$=2 \\ $d_2$=196\end{tabular}} 
   \\ \cline{3-8} 
 &
   &
  \begin{tabular}[c]{@{}c@{}}Inter-Level \\ Attention Layer\end{tabular} &
  \multicolumn{1}{c|}{\begin{tabular}[c]{@{}c@{}} N/A \end{tabular}} &
  \multicolumn{1}{c|}{\begin{tabular}[c]{@{}c@{}} N/A \end{tabular}} &
  \multicolumn{1}{c|}{\begin{tabular}[c]{@{}c@{}}$\alpha_2, \beta_2$=0.5,0.5 \\ $s_2$=1 \\ $p_2$=4 \\ $H_2$=1 \\ $E_2$=4 \end{tabular}} &
  \multicolumn{1}{c|}{\begin{tabular}[c]{@{}c@{}}$\alpha_2, \beta_2$=0.5,0.5 \\ $s_2$=1 \\ $p_2$=4 \\ $H_2$=1 \\ $E_2$=4 \end{tabular}} &
  \multicolumn{1}{c|}{\begin{tabular}[c]{@{}c@{}}$\alpha_2, \beta_2$=0.5,0.5 \\ $s_2$=1 \\ $p_2$=4 \\ $H_2$=1 \\ $E_2$=2 \end{tabular}} 
   \\ \cline{3-8} 
 &
   &
  \begin{tabular}[c]{@{}c@{}}Self-Attention \\ Block\end{tabular} &
  \multicolumn{1}{c|}{\begin{tabular}[c]{@{}c@{}}$R_2$=4 \\ $H_2$=2 \\ $E_2$=4  \\ $N_2$=2 \end{tabular}} &
  \multicolumn{1}{c|}{\begin{tabular}[c]{@{}c@{}}$R_2$=4 \\ $H_2$=2 \\ $E_2$=4  \\ $N_2$=2 \end{tabular}} &
  \multicolumn{1}{c|}{\begin{tabular}[c]{@{}c@{}}$R_2$=4 \\ $H_2$=2 \\ $E_2$=4  \\ $N_2$=2 \end{tabular}} &
  \multicolumn{1}{c|}{\begin{tabular}[c]{@{}c@{}}$R_2$=4 \\ $H_2$=2 \\ $E_2$=4  \\ $N_2$=2 \end{tabular}} &
  \multicolumn{1}{c|}{\begin{tabular}[c]{@{}c@{}}$W_2$=7 \\ $H_2$=6 \\ $E_2$=4  \\ $N_2$=2 \end{tabular}}
   \\ \hline
\multirow{3}{*}{Stage 3} &
  \multirow{3}{*}{$\frac{H}{16} \times \frac{W}{16}$} &
  Patch Merging &
  \multicolumn{1}{c|}{\begin{tabular}[c]{@{}c@{}}$K_3$=3 \\ $S_3$=2 \\ $d_3$=160\end{tabular}} &
  \multicolumn{1}{c|}{\begin{tabular}[c]{@{}c@{}}$K_3$=3 \\ $S_3$=2 \\ $d_3$=320\end{tabular}} &
  \multicolumn{1}{c|}{\begin{tabular}[c]{@{}c@{}}$K_3$=3 \\ $S_3$=2 \\ $d_3$=320\end{tabular}} &
  \multicolumn{1}{c|}{\begin{tabular}[c]{@{}c@{}}$K_3$=3 \\ $S_3$=2 \\ $d_3$=320\end{tabular}} &
  \multicolumn{1}{c|}{\begin{tabular}[c]{@{}c@{}}$K_3$=2 \\ $S_3$=2 \\ $d_2$=384\end{tabular}} 
   \\ \cline{3-8} 
 &
   &
  \begin{tabular}[c]{@{}c@{}}Inter-Level \\ Attention Layer\end{tabular} &
  \multicolumn{1}{c|}{\begin{tabular}[c]{@{}c@{}}$\alpha_3, \beta_3$=0.5,0.5 \\ $s_3$=1 \\ $p_3$=4 \\ $H_3$=1 \\ $E_3$=4 \end{tabular}} &
  \multicolumn{1}{c|}{\begin{tabular}[c]{@{}c@{}}$\alpha_3, \beta_3$=0.5,0.5 \\ $s_3$=1 \\ $p_3$=4 \\ $H_3$=1 \\ $E_3$=4 \end{tabular}} &
  \multicolumn{1}{c|}{\begin{tabular}[c]{@{}c@{}}$\alpha_3, \beta_3$=0.5,0.5 \\ $s_3$=1 \\ $p_3$=4 \\ $H_3$=1 \\ $E_3$=4 \end{tabular}} &
  \multicolumn{1}{c|}{\begin{tabular}[c]{@{}c@{}}$\alpha_3, \beta_3$=0.5,0.5 \\ $s_3$=3 \\ $p_3$=4 \\ $H_3$=1 \\ $E_3$=4 \end{tabular}} &
  \multicolumn{1}{c|}{\begin{tabular}[c]{@{}c@{}}$\alpha_3, \beta_3$=0.5,0.5 \\ $s_3$=2 \\ $p_3$=4 \\ $H_3$=1 \\ $E_3$=2 \end{tabular}} 
   \\ \cline{3-8} 
 &
   &
  \begin{tabular}[c]{@{}c@{}}Self-Attention \\ Block\end{tabular} &
  \multicolumn{1}{c|}{\begin{tabular}[c]{@{}c@{}}$R_3$=2 \\ $H_3$=5 \\ $E_3$=4  \\ $N_3$=2 \end{tabular}} &
  \multicolumn{1}{c|}{\begin{tabular}[c]{@{}c@{}}$R_3$=2 \\ $H_3$=5 \\ $E_3$=4  \\ $N_3$=2 \end{tabular}} &
  \multicolumn{1}{c|}{\begin{tabular}[c]{@{}c@{}}$R_3$=2 \\ $H_3$=5 \\ $E_3$=4  \\ $N_3$=2 \end{tabular}} &
  \multicolumn{1}{c|}{\begin{tabular}[c]{@{}c@{}}$R_3$=2 \\ $H_3$=5 \\ $E_3$=4  \\ $N_3$=6 \end{tabular}} &
  \multicolumn{1}{c|}{\begin{tabular}[c]{@{}c@{}}$W_3$=7 \\ $H_3$=12 \\ $E_3$=4  \\ $N_3$=6 \end{tabular}}
   \\ \hline
\multirow{3}{*}{Stage 4} &
  \multirow{3}{*}{$\frac{H}{32} \times \frac{W}{32}$} &
  Patch Merging &
  \multicolumn{1}{c|}{\begin{tabular}[c]{@{}c@{}}$K_4$=3 \\ $S_4$=2 \\ $d_4$=256\end{tabular}} &
  \multicolumn{1}{c|}{\begin{tabular}[c]{@{}c@{}}$K_4$=3 \\ $S_4$=2 \\ $d_4$=512\end{tabular}} &
  \multicolumn{1}{c|}{\begin{tabular}[c]{@{}c@{}}$K_4$=3 \\ $S_4$=2 \\ $d_4$=512\end{tabular}} &
  \multicolumn{1}{c|}{\begin{tabular}[c]{@{}c@{}}$K_4$=3 \\ $S_4$=2 \\ $d_4$=512\end{tabular}} &
  \multicolumn{1}{c|}{\begin{tabular}[c]{@{}c@{}}$K_4$=2 \\ $S_4$=2 \\ $d_4$=768\end{tabular}} 
   \\ \cline{3-8} 
 &
   &
  \begin{tabular}[c]{@{}c@{}}Inter-Level \\ Attention Layer\end{tabular} &
  \multicolumn{1}{c|}{\begin{tabular}[c]{@{}c@{}} N/A \end{tabular}} &
  \multicolumn{1}{c|}{\begin{tabular}[c]{@{}c@{}} N/A \end{tabular}} &
  \multicolumn{1}{c|}{\begin{tabular}[c]{@{}c@{}}$\alpha_4, \beta_4$=0.5,0.5 \\ $s_4$=1 \\ $p_4$=4 \\ $H_4$=1 \\ $E_4$=4 \end{tabular}} &
  \multicolumn{1}{c|}{\begin{tabular}[c]{@{}c@{}}$\alpha_4, \beta_4$=0.5,0.5 \\ $s_4$=1 \\ $p_4$=4 \\ $H_4$=1 \\ $E_4$=4 \end{tabular}} &
  \multicolumn{1}{c|}{\begin{tabular}[c]{@{}c@{}}$\alpha_4, \beta_4$=0.5,0.5 \\ $s_4$=1 \\ $p_4$=4 \\ $H_4$=1 \\ $E_4$=2 \end{tabular}} 
   \\ \cline{3-8} 
 &
   &
  \begin{tabular}[c]{@{}c@{}}Self-Attention \\ Block\end{tabular} &
  \multicolumn{1}{c|}{\begin{tabular}[c]{@{}c@{}}$R_4$=1 \\ $H_4$=8 \\ $E_4$=4  \\ $N_4$=2 \end{tabular}} &
  \multicolumn{1}{c|}{\begin{tabular}[c]{@{}c@{}}$R_4$=1 \\ $H_4$=8 \\ $E_4$=4  \\ $N_4$=2 \end{tabular}} &
  \multicolumn{1}{c|}{\begin{tabular}[c]{@{}c@{}}$R_4$=1 \\ $H_4$=8 \\ $E_4$=4  \\ $N_4$=2 \end{tabular}} &
  \multicolumn{1}{c|}{\begin{tabular}[c]{@{}c@{}}$R_4$=1 \\ $H_4$=8 \\ $E_4$=4  \\ $N_4$=2 \end{tabular}} &
  \multicolumn{1}{c|}{\begin{tabular}[c]{@{}c@{}}$W_4$=7 \\ $H_4$=24 \\ $E_4$=4  \\ $N_4$=2 \end{tabular}}
   \\ \hline
\end{tabular}
\end{adjustbox}
\label{table:hyperparameters}
\end{center}
\end{table}

\subsection{Source of Baseline Results}

We detail the sources of baseline results in our main results table. We use two main sources for the results in our baselines: the mmsegmentation repository \cite{mmseg2020} and from past papers.

\paragraph{Mmsegmentation} We use implementations in the mmsegmentation repository for FCN \cite{long2015fcn}, PSPNet \cite{zhao2017pspnet}, UperNet \cite{xiao2018upernet}, CCNet \cite{huang2019ccnet}, DANet \cite{fu2019danet} and Swin \cite{hvt:liu2021Swin}. For each, we take the pretrained model with the best reported results and use this model to evaluate our mIOU and F-Score metric. In the case of Swin Transformer, we report the performance of the mmsegmentation implementation instead of the original paper as we build \method{} on top of this model. This facilitates a clearer comparison of the improvements that our method brings.   

\paragraph{Past Papers} We report numbers for OCRNet \cite{yuan2020ocrnet} and Segformer \cite{hvt:xie2021segformer} from the Segformer results, and Gated-SCNN \cite{takikawa2019gatedscnn} and D3Net \cite{takahashi2021d3net} from their respective papers. We implement \method{} on Segformer's codebase and report mIOU and F1 using pretrained models provided by the authors.

\subsection{Modified ADE20K F-Score Calculation}

The main goal of our modified F-score is to differentiate our output's effect to delineate borders from the classification accuracy. The normal F-score defined by Perazzi et. al~\cite{davisPerazzi2016} is a precision/recall metric calculated on each class separately. Given a specific class, we use the contour operation $c(\cdot)$ to obtain the contours of the output mask $c(M_{\text{class}})$ and the ground truth mask $c(G_{\text{class}})$. The F-score is then calculated using the precision $P_{\text{class}}$ and recall $R_{\text{class}}$ between these two contours across all classes:

\begin{equation}
\mathcal{F} = \frac{1}{N_{classes}}\sum_{class}\frac{2P_{\text{class}}R_{\text{class}}}{P_{\text{class}}+ R_{\text{class}}}
\end{equation} 

In ADE20K this equation is not ideal as there are 150 semantic categories. As a result, often the contour is predicted correctly but the class is not, resulting in a F-score of 0. In addition, classes which are not present in an image are assigned a score of 1 automatically. Over the whole test dataset, this results in very high numbers with low variance due to the small subset of classes an image actually contains. To remedy these issues, we take the image-wise F-score. Instead of computing across each individual class, we combine the contours of each class into a single image-wise contour. By doing this, we disambiguate classification accuracy from the contour accuracy which we are interesting in evaluating:

\begin{equation}
\mathcal{F}_{mod} = \frac{2P_{\text{image}}R_{\text{image}}}{P_{\text{image}}+ R_{\text{image}}}
\end{equation} 

Table \ref{table:f1modcomparison} shows the difference in scores between both metrics. We can see that our modified score is much more expressive, allowing for better disambiguation between the effectiveness of different models on the ADE20K dataset. 

\begin{table}[]
\begin{center}
\caption{\textbf{Comparison between Normal F-Score and Modified F-Score}}
\begin{adjustbox}{width=0.5\textwidth}
\begin{tabular}{|c|c|c|}
\hline
\textbf{Model} &\textbf{Normal F-Score} & \textbf{Modified F-Score (Ours)} \\ \hline
SegFormer-B0 &   88.2 &    67.2                \\ \hline
+ \methodone{} &  88.3  &   69.4                 \\ \hline
SegFormer-B1 & 88.1   &      70.6              \\ \hline
+ \methodone{} &  88.2  &      74.1              \\ \hline
\end{tabular}
\end{adjustbox}
\label{table:f1modcomparison}
\end{center}
\end{table}

\section{Further Ablations}

We explore further ablations for each HILA-specific hyper-parameter in Table~\ref{table:hyperparameters}.

\label{sec:more_ablations}




\begin{table}[]
\begin{center}
\caption{\footnotesize \textbf{Further Ablation Studies of Our Method on Cityscapes.} a) We ablate the effects of our information propagation constant on the SegFormer-B0 + \methodone{} Model, b) We ablate the effects of sharing the weights of \method{} components across iterations on the SegFormer-B0 + \methodone{} Model, c) We ablate between different strides in applying \method{} on the SegFormer-B2 + \methodthree{} Model, d) We test different sizes of local patches on the SegFormer-B1 + \methodthree{} Model.}
\begin{subtable}{.5\linewidth}
\caption{\footnotesize Information Propagation Constant}
\vspace{-4mm}
\begin{center}
\label{table:appendix_informationpropagation}
\begin{adjustbox}{width=0.5\textwidth}
\begin{tabular}{|c|c|c|}
\hline
$\alpha, \beta$ & \begin{tabular}[c]{@{}c@{}}mIOU \\ (SS)\end{tabular} & \begin{tabular}[c]{@{}c@{}}F-Score\\ 3px (SS)\end{tabular} \\ \hline
(1.0, 1.0) &          76.9                &            70.4              \\
(0.2, 0.8) &          76.3                &             69.1             \\
(0.5, 0.5) &          \textbf{77.1}                &                 \textbf{70.5}         \\
(0.8, 0.2) &            76.5              &                 70.0         \\ \hline
\end{tabular}
\end{adjustbox}
\end{center}
\end{subtable}
\begin{subtable}{\linewidth}
\caption{\footnotesize \method{} Weight Sharing}
\vspace{-4mm}
\begin{center}
\begin{adjustbox}{width=0.5\textwidth}
\label{table:appendix_weightsharing}
\begin{tabular}{|c|c|c|c|c|}
\hline
Dataset & Weight Sharing & Params (M) & \begin{tabular}[c]{@{}c@{}}mIOU \\ (SS)\end{tabular} & \begin{tabular}[c]{@{}c@{}}F-Score\\ 3px (SS)\end{tabular} \\ \hline
ADE20K                       &     \xmark       &    4.4     & 39.3   &  \textbf{69.7}       \\ 
                       &     \cmark       &    4.2     & \textbf{40.3}   &  69.4
\\ 
Cityscapes                       &     \xmark      &    4.4     & 77.1   &  \textbf{70.5} 
\\ 
                       &     \cmark       &    4.2     & \textbf{77.3}   &  70.3       \\ \hline
\end{tabular}
\end{adjustbox}
\end{center}
\end{subtable}
\begin{subtable}{\linewidth}
\caption{\footnotesize \method{} Update Stride}
\vspace{-4mm}
\begin{center}
\begin{adjustbox}{width=0.5\textwidth}
\label{table:appendix_updatestride}
\begin{tabular}{|c|c|c|c|c|}
\hline
\begin{tabular}[c]{@{}c@{}}Stage 3\\ Stride Size\end{tabular} & Params (M) & FLOPS ↓ & \begin{tabular}[c]{@{}c@{}}mIOU \\ (SS)\end{tabular} & \begin{tabular}[c]{@{}c@{}}F-Score\\ 3px (SS)\end{tabular} \\ \hline
1 & 30.8  & 956.8 & 80.8            & 74.4 \\
2 & 30.8  & 879.2 & 81.2           & \textbf{75.0} \\ 
3 &  30.8 &  \textbf{867.4} & \textbf{81.5} &    74.9                         \\ \hline
\end{tabular}
\end{adjustbox}
\end{center}
\end{subtable}
\begin{subtable}{\linewidth}
\caption{\footnotesize Local Patch Size}
\vspace{-4mm}
\begin{center}
\begin{adjustbox}{width=0.5\textwidth}
\label{table:appendix_patchsize}
\begin{tabular}{|c|c|c|c|c|}
\hline
Patch Size & Params (M) & FLOPS ↓ & \begin{tabular}[c]{@{}c@{}}mIOU \\ (SS)\end{tabular} & \begin{tabular}[c]{@{}c@{}}F-Score\\ 3px (SS)\end{tabular} \\ \hline
2                       &     21.6       &    417.5     & 80.7   &  \textbf{74.7}       \\ 
4                       &     21.6       &    417.8     & \textbf{80.8}   &  74.5       \\ 
6                       &     21.6       &    418.3     & 80.8   &  74.5       \\ \hline
\end{tabular}
\end{adjustbox}
\end{center}
\end{subtable}
\end{center}
\end{table}

\begin{figure*}[t!]
\centering
\includegraphics[width=0.9\linewidth, trim=0cm 63cm 30cm 0cm, clip]{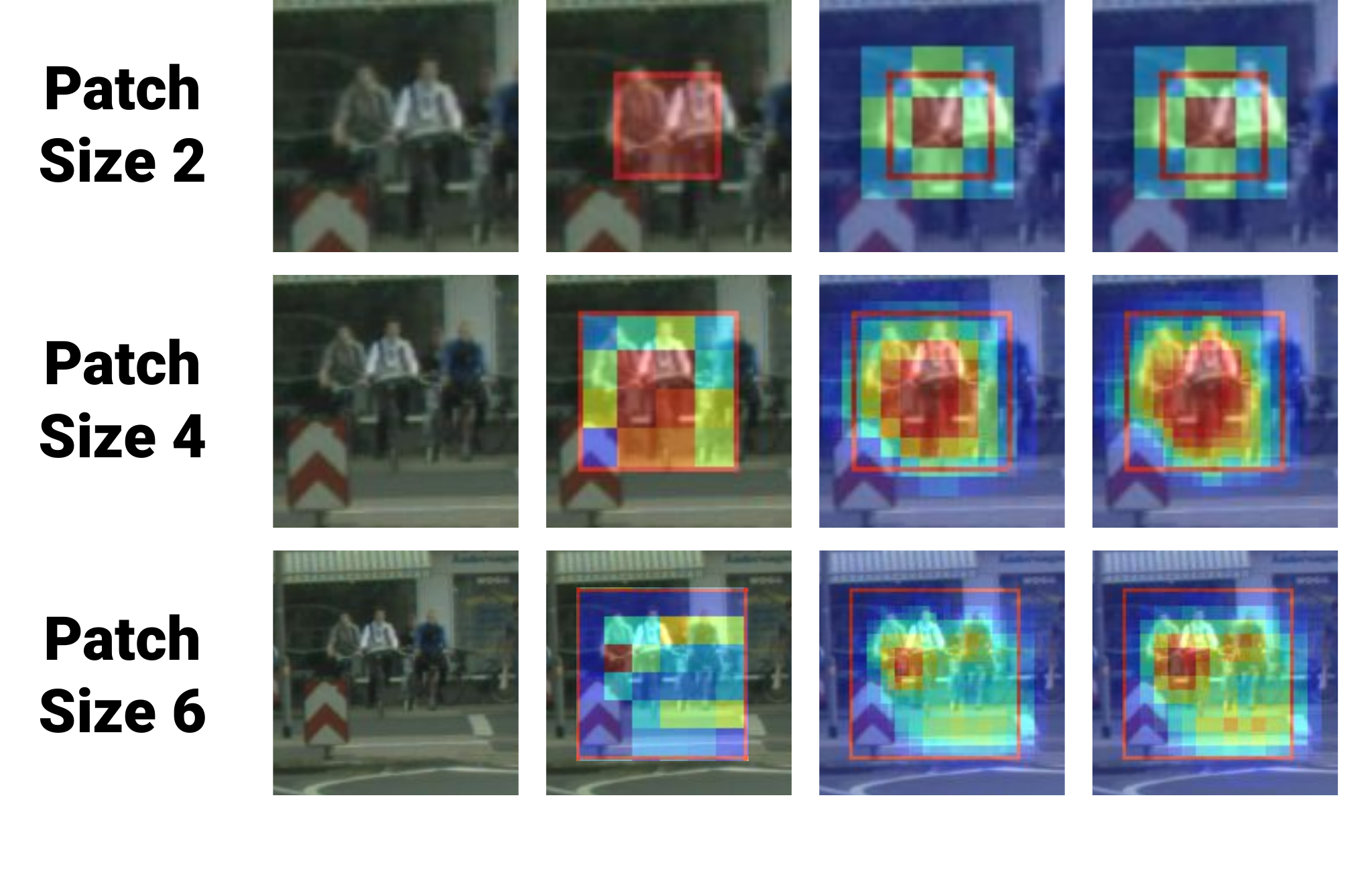} 
\vspace{-7mm}
\caption{\footnotesize \textbf{Attention Across Patch Sizes.} The second column shows Stage 4 to 3 attention, the third column shows Stage 4 to 2 attention, and the fourth column shows Stage 4 to Stage 1. We obtain the best attention results at $p_{\ell} = 4$. At $p_{\ell} = 2$, our attention is uniform across all patches. At $p_{\ell} = 6$, the larger local area results in a larger variety of objects being mixed into the attention.}
\label{figure:patchsizeattention}
\vspace{-4mm}
\end{figure*}

\subsection{Patch Size Ablation}

We ablate the patch size of our Inter-Level attention, $p_{\ell}$. We ablate between patch sizes of 2, 4, and 6 on our SegFormer B1 + \methodthree{} model and show our results in Table~\ref{table:appendix_patchsize}. Our ablation results show that we obtain similarly good results at all patch sizes when adding \method{} to a model. Intuitively, there is a strong inductive bias that the center-most features in a local-patch are most likely to be related to it's corresponding higher-level feature. Due to this, \method{}'s inter-level update propagates meaningful information, even in the smallest patch size of 2. In the case of larger patch sizes, such as 6, there are more overlapping higher-level features for each lower-level feature. This results in a more complex relationship that may be harder to learn - resulting in results very similar to smaller patch parameters but with higher compute required. We use the local patch size of 4, which has the best trade-off between complexity, performance, meaningful attention visualizations. We show examples of our attention in different patch sizes in Figure~\ref{figure:patchsizeattention}. We see the best attention results at $p_{\ell} = 4$. At $p_{\ell} = 2$, our attention is the same across all patches due to having no overlaps between local patches between higher-level features. At $p_{\ell} = 6$, our larger local area results in less focused attention, with multiple objects being attended on. 

\subsection{\method{} Weight Sharing}

We ablate the effects of sharing the weights of \method{} components across iterations and show our results in Table~\ref{table:appendix_weightsharing}. Our ablation shows that we obtain similar results when sharing weights and when not sharing weights. In the case that we share weights in our \method{} components, we obtain higher mIOU and in the case where we do not share weights, we obtain higher F-scores. We decide to share the weights of our \method{} components between iterations, as this allows us to reduce the number of parameters needed. This is particularly significant in larger models, where there are many iterations per stage. 

\subsection{\method{} Stride}

We ablate the stride at which \method{} is applied in larger models with more iterations. Two of our models have stride $s_\ell > 1$: SegFormer B2 + \methodthree{} and Swin-T + \methodthree{}. In both cases, this occurs in Stage 3, where both models have 6 blocks. We use our SegFormer B2 + \methodthree{} model to ablate our stride and show our results in Table~\ref{table:appendix_updatestride}. Both stride values of 2 and 3 provide benefits in performance while being significantly lighter in computation. For SegFormer B2, we use a stride of 3 in Stage 3, resulting in \method{} replacing 2 blocks total instead of 6 blocks total. For the Swin-T + \methodthree{}, we find that a stride of 2 provides better results. Our ablations suggest it is unnecessary to replace every block with \method{} - intuitively, a higher stride in applying \method{} allows the model to refine higher-level information in more detail before propagating this information between levels in the \method{} block. 

\subsection{Information Propagation Constant}
We ablate the information propagation constants, $\alpha$ and $\beta$, that we use in the FFN of our Inter-Level Attention. For this ablation, we compare different constants in our B0 + \methodone{} model. We compare against the baseline case of residual addition (1.0, 1.0), as well as other constants of (0.2, 0.8) and (0.8, 0.2), and show our results in Table~\ref{table:appendix_informationpropagation}. We find that across various information propagation constants, (0.5, 0.5) results in the best result of 77.1 mIOU and 70.5 F-Score. We use (0.5, 0.5) for all of our models across all stages, as we find that it creates the best results across the majority of settings. 

In our initial experiments on a play dataset, we find that the information propagation constants result in greater improvements in feature strength across iterations. For these play experiments, we train several SegFormer B2 + \methodthree{} models with different information propagation constants on the CelebAMask-HQ Dataset~\cite{CelebAMask-HQ} at a 256x256 resolution. We follow the same training configuration as outlined in the main paper, with two changes being that we do not use Imagenet1K pre-training, and we use a smaller decoding head with 128 hidden channels due to the smaller size of the play dataset. We evaluate the strength of our intermediate features using linear probing with the mIOU metric. We freeze our model weights, and train a separate decoding head for each feature at all stages over all iterations. These individual decoding heads follows the same design as the main decoding head. By doing so, we can evaluate our features by comparing and contrast the mIOU improvement in our features due to the inter-level information passed in our \method{} blocks.

\begin{figure*}[t!]
\centering
\includegraphics[width=\linewidth, trim=0cm 3cm 0cm 0cm, clip]{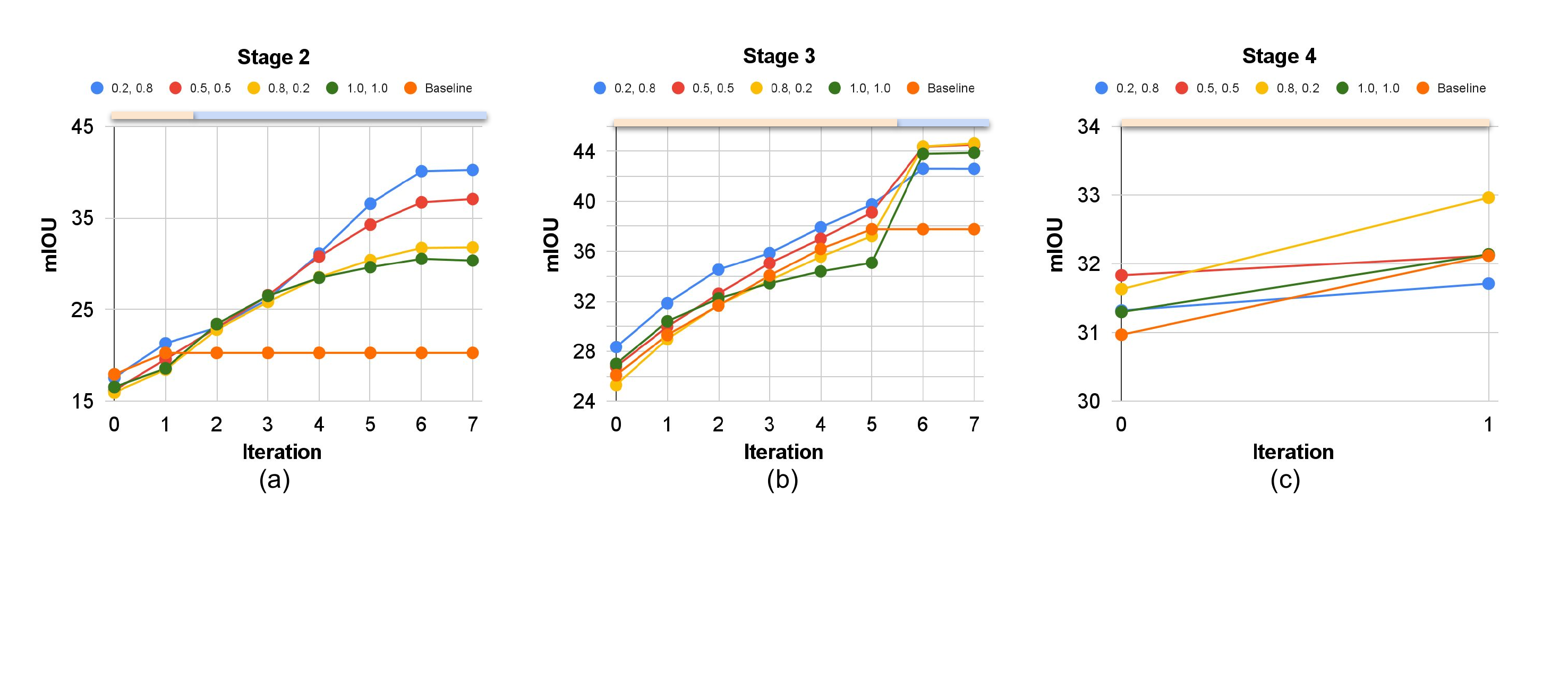} 
\vspace{-7mm}
\caption{\footnotesize \textbf{Linear Probing of Features.} We train linear probing heads for features at each iteration and evaluate the mIOU to show the improvement in feature strength. In a) we show Stage 2 features, in b) we show Stage 3 features, and in c) we show Stage 4 features. We can see that the inter-level information passed by \method{} results in vastly improved mIOU on the features as compared to the baseline (orange). In addition, by adding the information propagation constants, this improvement is further improved over the baseline of (1.0, 1.0) which is the residual update (green).}
\label{figure:linearprobing}
\vspace{-4mm}
\end{figure*}

As shown in Figure~\ref{figure:linearprobing}, adding \method{} to SegFormer results in improved feature quality as compared to the baseline SegFormer model. We can see that extra iterations where inter-level information is passed down through the \method{} block significantly improves feature quality above the baseline. In addition, adding information propagation constants improves the amount of improvement. As compared to the residual case with weights of $\alpha, \beta = (1.0, 1.0)$, having momentum-like weights where the values of $\alpha$ and $\beta$ add to 1 results in larger mIOU increase in our features across iterations. We can observe that the linear probing charts show that the weights of $\alpha, \beta = (0.5, 0.5)$ have the best balance of improvement across all feature stages, and this is supported by our ablation experiments on Cityscapes in Table~\ref{table:appendix_informationpropagation}. We theorize that having momentum-like weights allow for larger amounts of information in our features to be 'replaced' by different level information passed down/up through the \method{} block as compared to the residual case, allowing for more improvement.

\clearpage
\section{More Segmentation Results}
\label{sec:more_seg_results}
We provide extra qualitative segmentation results. The evaluation setting follows the main paper, and we compare the baseline SegFormer B1 model and SegFormer B1 with \methodthree{}.

\begin{figure*}[t!]
\centering
\vspace{-5mm}
\includegraphics[width=\linewidth, trim=0cm 6.0cm 0cm 0cm, clip]{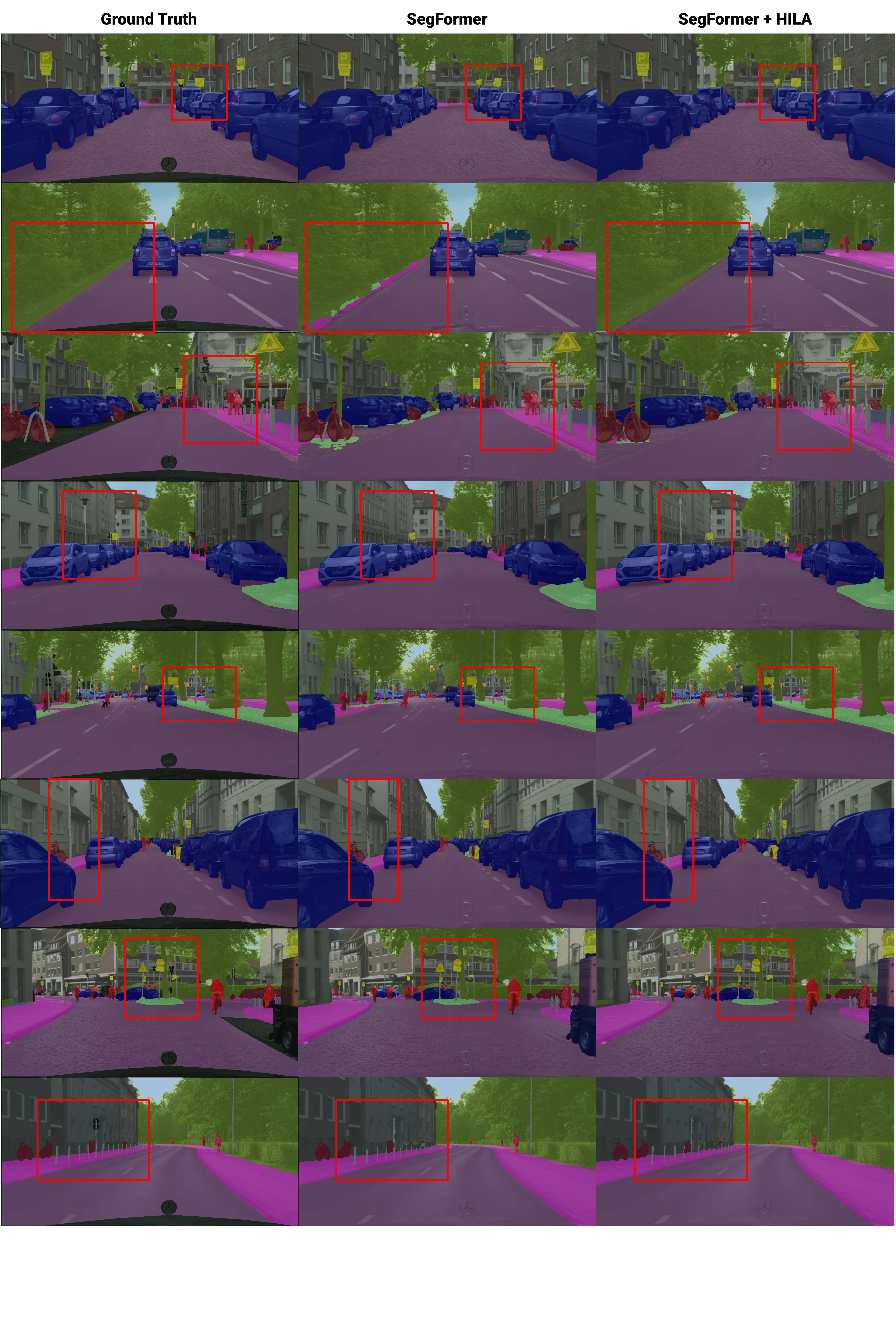} 
\vspace{-7mm}
\caption{\footnotesize \textbf{More Qualitative comparisons on Cityscapes.} We compare our method with the SegFormer baseline and highlight the differences in red rectangles. Once again, we see significantly better classification and boundaries, especially in narrow or ambiguous objects. }
\label{figure:qualitativeevaluation2}
\vspace{-4mm}
\end{figure*}

\begin{figure*}[t!]
\centering
\vspace{-5mm}
\includegraphics[width=\linewidth, trim=0cm 5cm 0cm 0cm, clip]{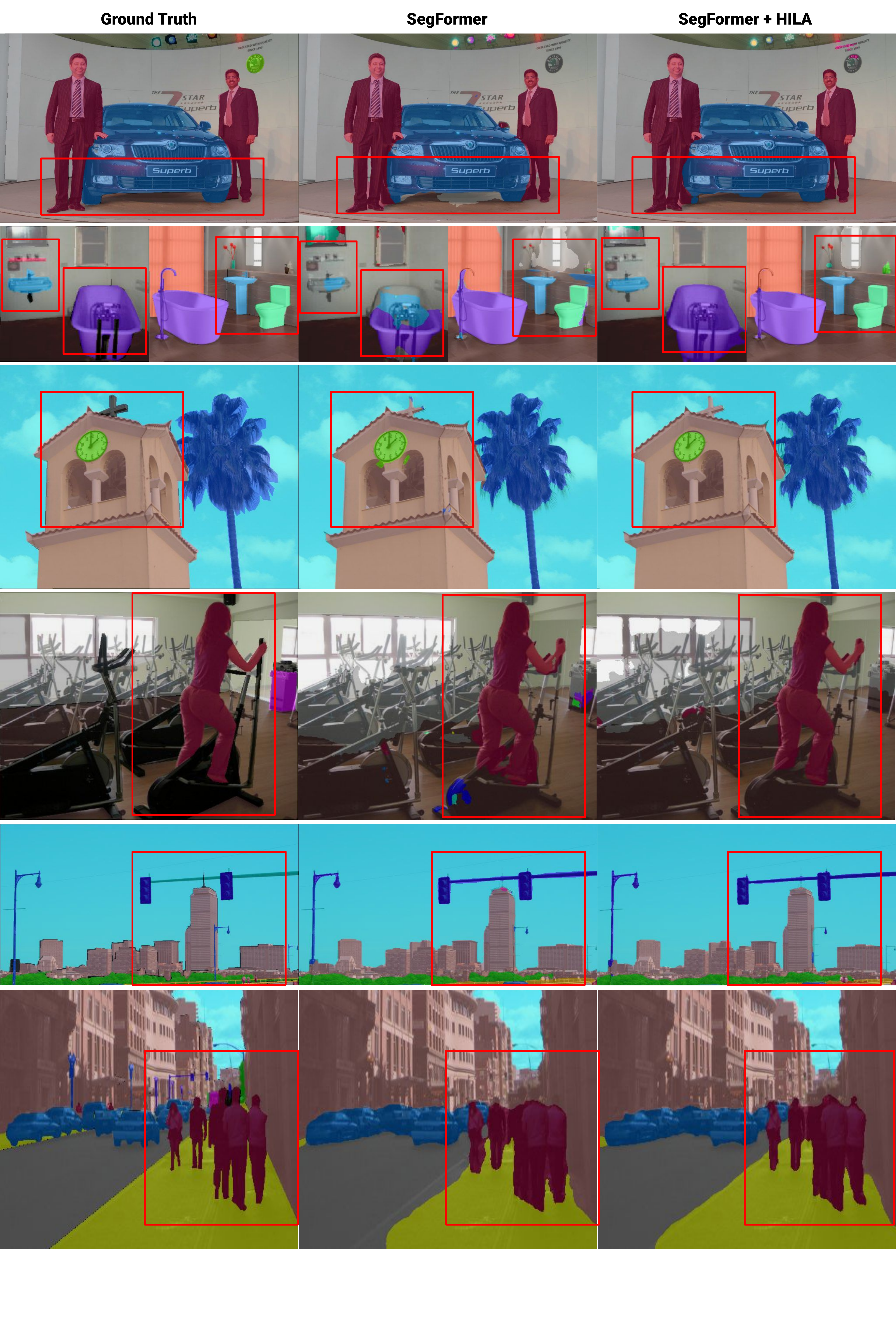} 
\vspace{-7mm}
\caption{\footnotesize \textbf{Qualitative comparisons on ADE20K.} We compare our method with the SegFormer baseline and highlight the differences using red rectangles. Our predictions conform to boundaries object boundaries much better, and in some cases, surpasses the coarse ground truth masks.}
\label{figure:qualitativeevaluation_ADE}
\vspace{-4mm}
\end{figure*}

\begin{figure*}[t!]
\centering
\vspace{-5mm}
\includegraphics[width=\linewidth, trim=0cm 0cm 0cm 0cm, clip]{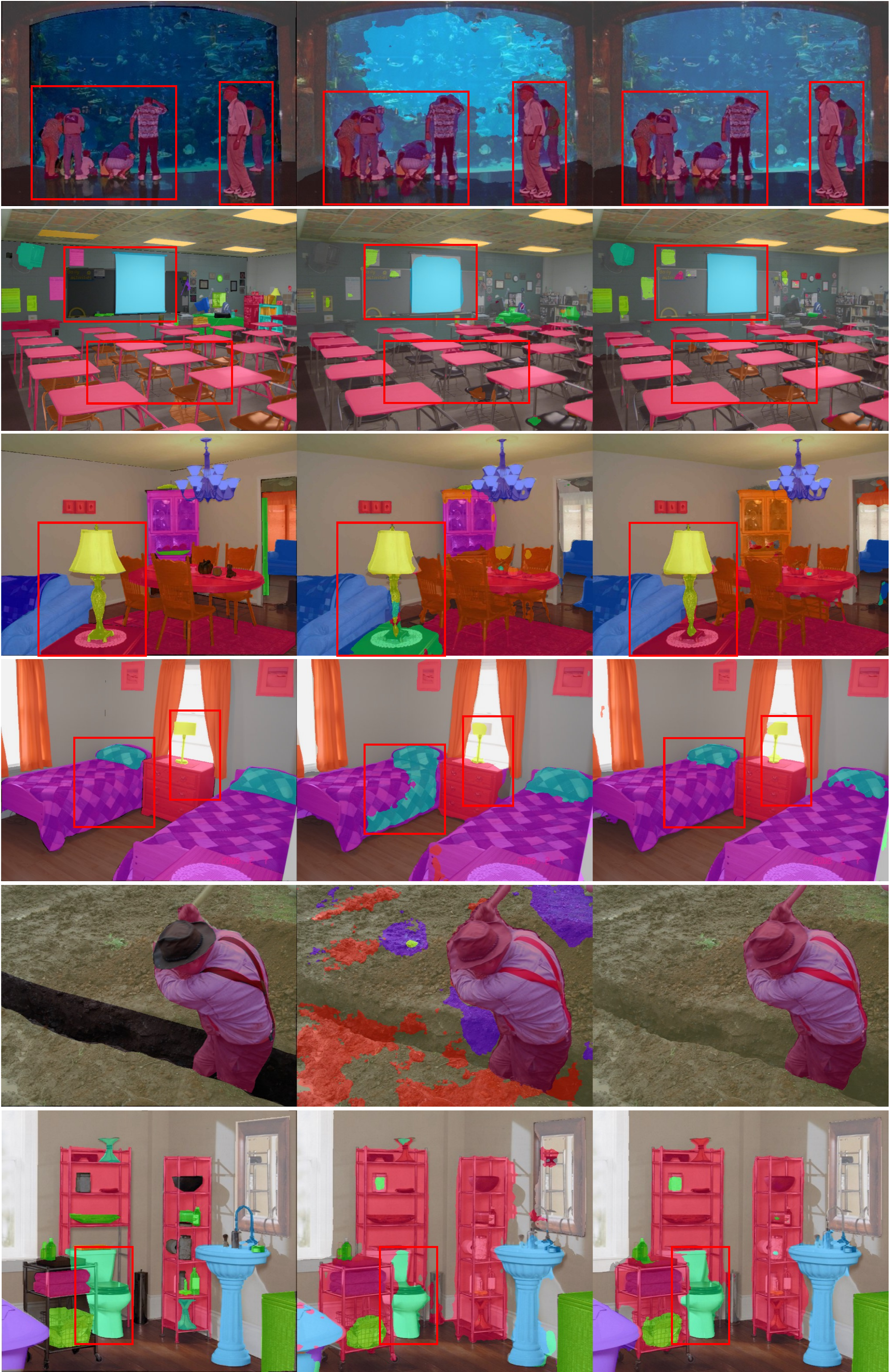} 
\vspace{-7mm}
\caption{\footnotesize \textbf{Qualitative comparisons on ADE20K.} We compare our method with the SegFormer baseline and highlight the differences using red rectangles. Our predictions conform to boundaries object boundaries much better, and in some cases, surpasses the coarse ground truth masks.}
\label{figure:qualitativeevaluation_ADE2}
\vspace{-4mm}
\end{figure*}

\begin{figure*}[t!]
\centering
\vspace{-5mm}
\includegraphics[width=\linewidth, trim=0cm 35.0cm 0cm 0cm, clip]{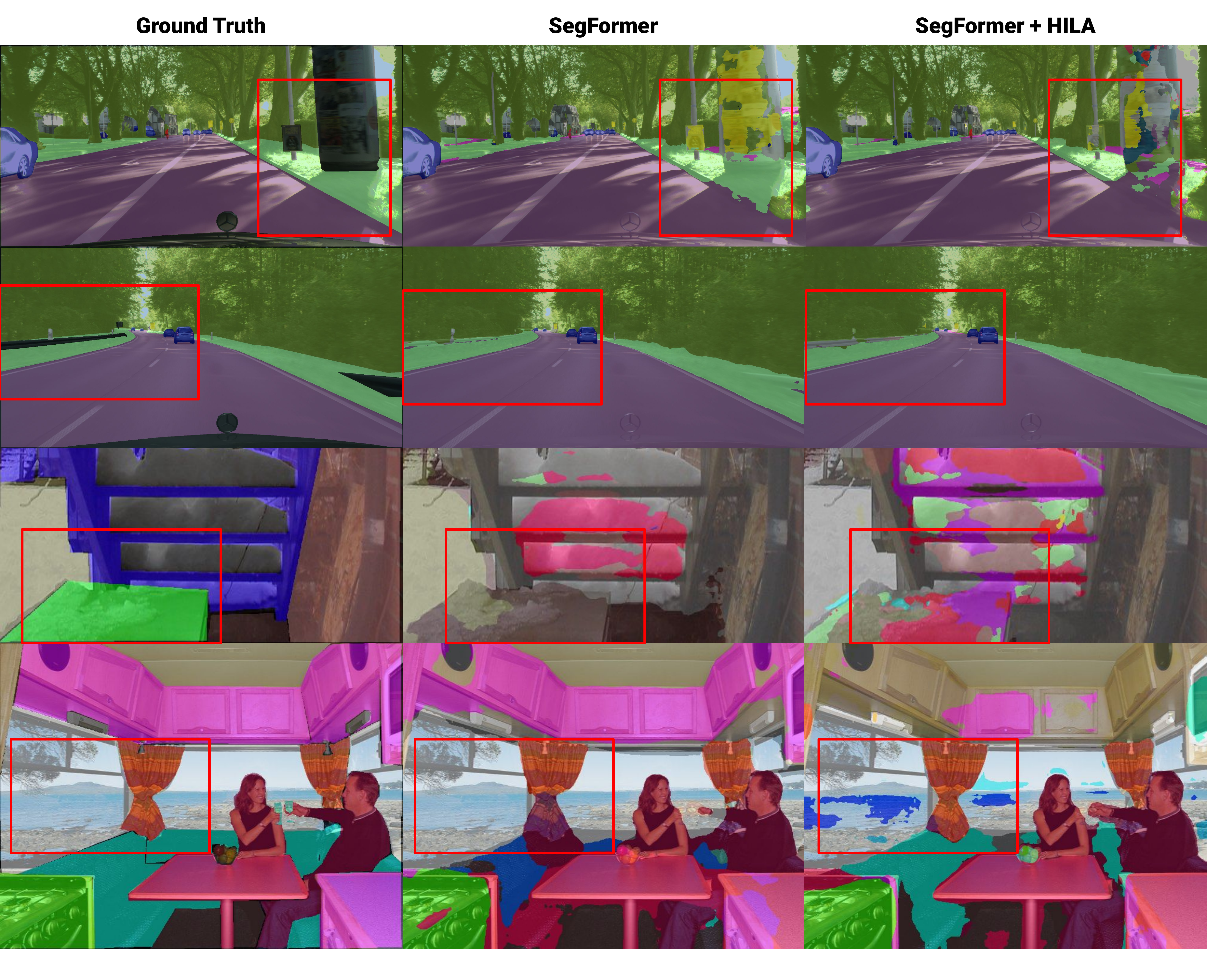} 
\vspace{-7mm}
\caption{\footnotesize \textbf{Example Failure Cases.} We show 2 cases out of our 10 worst segmentations compared to the baseline SegFormer model in both Cityscapes and ADE Validation set. In Cityscapes, our worst failure cases are mainly due to improper segmentation of large-sized background classes. In ADE20K, our worst failure cases tends to be when confusion in the higher-level features leads to incorrectly hallucinating a wrong class due to the large amount of semantic classes in the dataset.}
\label{figure:qualitativeevaluation_failure}
\vspace{-4mm}
\end{figure*}

We provide additional qulitative results comparing with SegFormer B1 on Cityscapes in Figure~\ref{figure:qualitativeevaluation} and Figure~\ref{figure:qualitativeevaluation2}. More qualitive results for ADE20K are shown in Figure~\ref{figure:qualitativeevaluation_ADE} and Figure~\ref{figure:qualitativeevaluation_ADE2}, with some failure cases in Figure~\ref{figure:qualitativeevaluation_failure}.
We outperform the baselines, with improvements particularly evident in boundaries for narrow and ambiguous objects. 

\section{More Hierarchichal Attention Results}
\label{sec:more_attn_results}
We provide extra qualitative hierarchichal attention results. The evaluation setting follows the main paper, with our attention coming from the SegFormer B1 with \methodthree{} model.

\begin{figure*}[t!]
\centering
\vspace{-5mm}
\includegraphics[width=\linewidth, trim=0cm 46.0cm 28cm 0cm, clip]{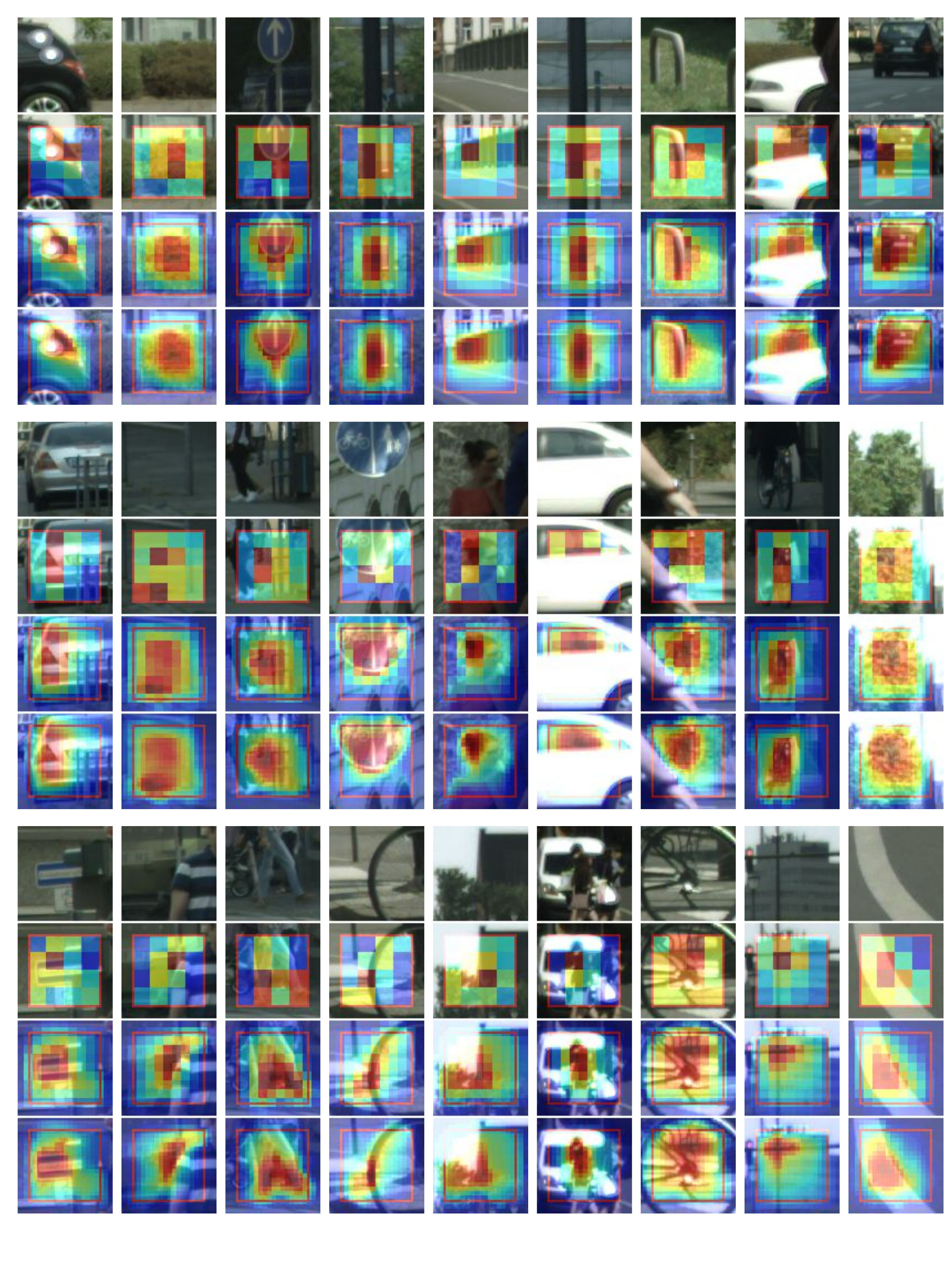} 
\vspace{-7mm}
\caption{\footnotesize \textbf{Visualization of Hierarchical Attention on Cityscapes.} We show the original image in the top row, and our hierarchical attention with each row adding attention from a lower-level stage. The 2nd row shows attention from Stage 4,  the 3rd row shows Stage 3\&4, and the last row shows Stage 2\&3\&4. The red box represents receptive field for the Stage 4 features. Our attention mask captures the object boundary from higher-level to lower-level, showing the hierarchy from the whole object into finely-detailed part mask We capture thin objects, such as poles, object boundaries for complex objects, and the background.}
\label{figure:cityscapes_attention}
\vspace{-4mm}
\end{figure*}

\begin{figure}[t!]
\vspace{-5mm}
\centering
\includegraphics[width=0.1\textwidth]{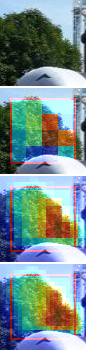}
\includegraphics[width=0.1\textwidth]{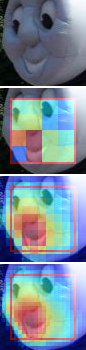}
\includegraphics[width=0.1\textwidth]{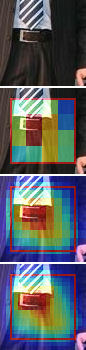}
\includegraphics[width=0.1\textwidth]{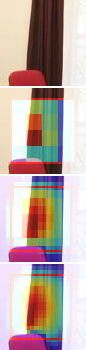}
\includegraphics[width=0.1\textwidth]{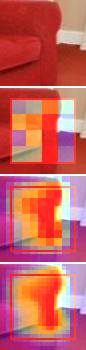}
\includegraphics[width=0.1\textwidth]{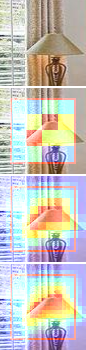}
\includegraphics[width=0.1\textwidth]{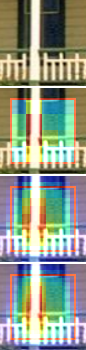}
\includegraphics[width=0.1\textwidth]{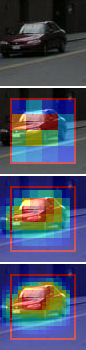}
\includegraphics[width=0.1\textwidth]{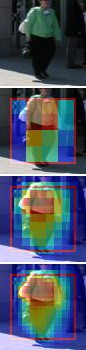}
\includegraphics[width=0.1\textwidth]{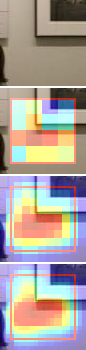}
\includegraphics[width=0.1\textwidth]{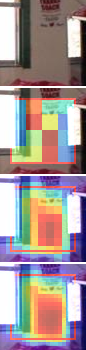}
\includegraphics[width=0.1\textwidth]{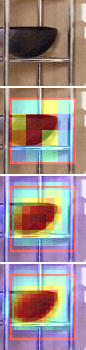}
\includegraphics[width=0.1\textwidth]{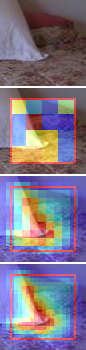}
\includegraphics[width=0.1\textwidth]{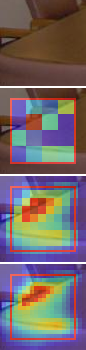}
\includegraphics[width=0.1\textwidth]{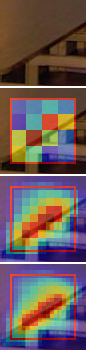}
\includegraphics[width=0.1\textwidth]{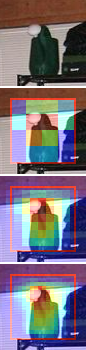}
\includegraphics[width=0.1\textwidth]{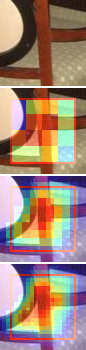}
\includegraphics[width=0.1\textwidth]{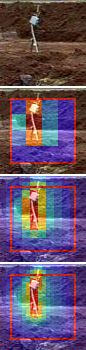}
\includegraphics[width=0.1\textwidth]{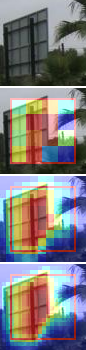}
\includegraphics[width=0.1\textwidth]{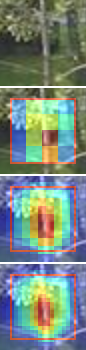}
\includegraphics[width=0.1\textwidth]{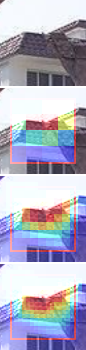}
\includegraphics[width=0.1\textwidth]{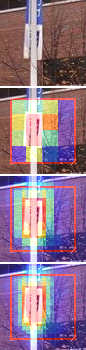}
\includegraphics[width=0.1\textwidth]{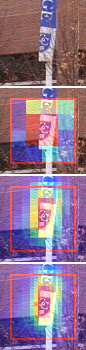}
\includegraphics[width=0.1\textwidth]{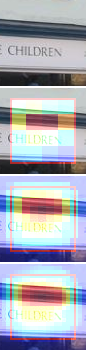}
\includegraphics[width=0.1\textwidth]{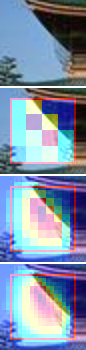}
\includegraphics[width=0.1\textwidth]{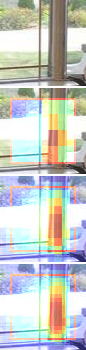}
\includegraphics[width=0.1\textwidth]{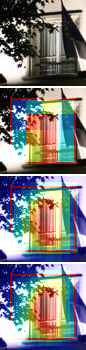}
\vspace{-3mm}
\caption{\footnotesize \textbf{Visualization of Hierarchical Attention on ADE.} We show the original image in the top row, and our hierarchical attention with each row adding attention from a lower-level stage. The 2nd row shows attention from Stage 4,  the 3rd row shows Stage 3\&4, and the last row shows Stage 2\&3\&4. The red box represents receptive field for the Stage 4 features.  Our attention mask captures the object boundary from higher-level to lower-level, showing the hierarchy from the whole object into finely-detailed part mask We capture thin objects, such as poles, object boundaries for complex objects, and the background.}
\label{figure:ade_attention}
\vspace{-5mm}
\end{figure}

We show additional qualitative results with hierarchichal attention on Cityscapes in Figure~\ref{figure:cityscapes_attention} and on ADE in Figure~\ref{figure:ade_attention}.

\begin{figure*}[t!]
\centering
\vspace{-5mm}
\includegraphics[width=\linewidth, trim=0cm 15.0cm 0cm 0cm, clip]{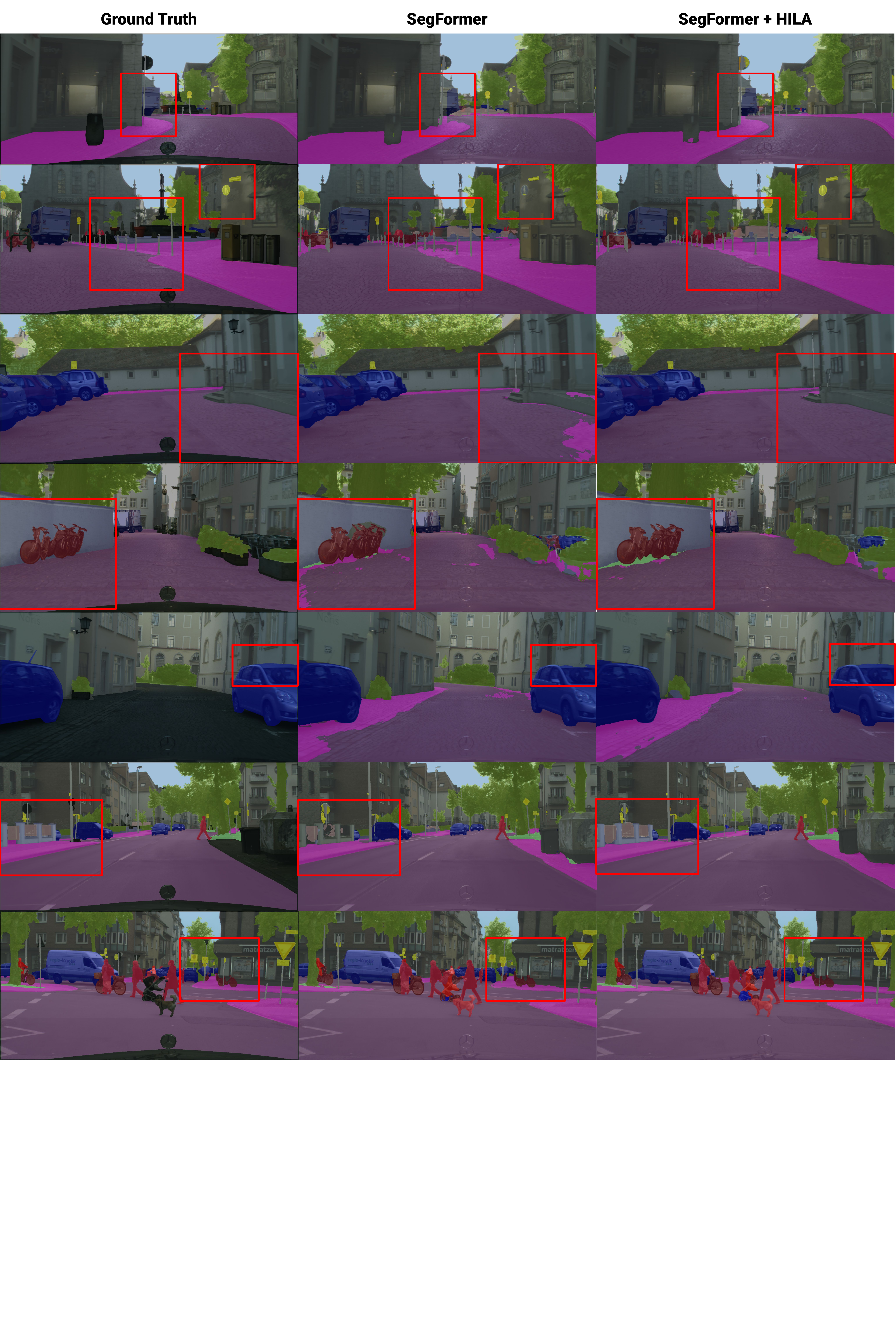} 
\vspace{-7mm}
\caption{\footnotesize \textbf{More Qualitative comparisons on Cityscapes.} We compare our method with the SegFormer baseline and highlight the differences in red rectangles. Once again, we see significantly better classification and boundaries, especially in narrow or ambiguous objects. }
\label{figure:qualitativeevaluation}
\vspace{-4mm}
\end{figure*}

\end{document}